\documentclass[letterpaper, 10 pt, conference]{ieeeconf}  

\IEEEoverridecommandlockouts                              

\usepackage{cite}
\usepackage{xcolor}

\usepackage{graphics} 
\usepackage{epsfig} 
\usepackage{mathptmx} 
\usepackage{amsmath} 
\usepackage{amssymb}  
\usepackage{algorithm}
\usepackage{algpseudocode}
\usepackage{color}
\usepackage{graphicx}
\usepackage{subcaption}
\usepackage{diagbox}
\newcommand{\dd}{\textrm{d}}

\title{\LARGE \bf
Developing Simulation Models for Soft Robotic Grippers in Webots}

\author{Yulyan Wahyu Hadi*$^{1,2}$, Lars Hof*$^{1}$, Bayu Jayawardhana$^{1}$, and Bahar Haghighat$^{1}$
\thanks{*Authors with equal contribution.}
\thanks{$^{1}$Y. Hadi, L. Hof, B. Jayawardhana, and B. Haghighat are all with the Faculty of Science and Engineering,
        University of Groningen, 9747 AG Groningen, The Netherlands
        {\tt\small y.w.hadi@rug.nl}, {\tt\small l.hof@student.rug.nl}, {\tt\small b.jayawardhana@rug.nl},  {\tt\small bahar.haghighat@rug.nl}}%
\thanks{$^{2}$Y. Hadi is also with the School of Electrical Engineering and Informatics, Bandung Institute of Technology, Indonesia}
}
\begin{document}

\maketitle
\thispagestyle{empty}
\pagestyle{empty}
\begin{abstract}
Robotic simulators provide cost-effective and risk-free virtual environments for studying robotic designs, control algorithms, and sensor integrations. They typically host extensive libraries of sensors and actuators that facilitate rapid prototyping and design evaluations in simulation. The use of the most prominent existing robotic simulators is however limited to simulation of rigid-link robots. On the other hand, there exist dedicated specialized environments for simulating soft robots. This separation limits the study of soft robotic systems, particularly in hybrid scenarios where soft and rigid sub-systems co-exist. In this work, we develop a lightweight open-source digital twin of a commercially available soft gripper, directly integrated within the robotic simulator Webots. We use a Rigid-Link-Discretization (RLD) model to simulate the soft gripper. 
Using a Particle Swarm Optimization (PSO) approach, we identify the parameters of the RLD model based on the kinematics and dynamics of the physical system and show the efficacy of our modeling approach in validation experiments. All software and experimental details are available on github: \textit{https://github.com/anonymousgituser1/Robosoft2025}.


\end{abstract}
\section{Introduction}
Soft robots are primarily constructed from highly flexible materials and often draw inspiration from the mechanical behavior of living organisms \cite{Mazzolai}. Researchers attempt to fabricate soft robots by mimicking the compliance and morphology of living creatures, with elastic bodies \cite{Rus} and compliant actuation \cite{Falkenhahn}, such as soft grippers \cite{Galloway}\cite{Elfferich}\cite{Abondance}, soft manipulator \cite{Laschi}, and soft robotic ﬁsh \cite{Katzschmann}. The inherent softness and low mechanical resistance of soft robots allow them to interact safely with delicate objects and adapt passively to complex surfaces. Multiple physics-based robotic simulators exist for traditional rigid-link robots, such as Gazebo \cite{Koenig} and Webots \cite{Michel}, which provide powerful tools for designing both the hardware and control of these robots, including integration with various kinds of virtual sensors, actuators, and other simulated robots. However, these do not support the simulation of soft robots, limiting the potential for systematic design and development in this field. This also limits the study of hybrid systems of soft and rigid robots within the same simulation environment. 

\begin{figure}[t]
   \centering
    \begin{subfigure}[t]{0.49\linewidth}
        \centering
        \includegraphics[width=\linewidth]{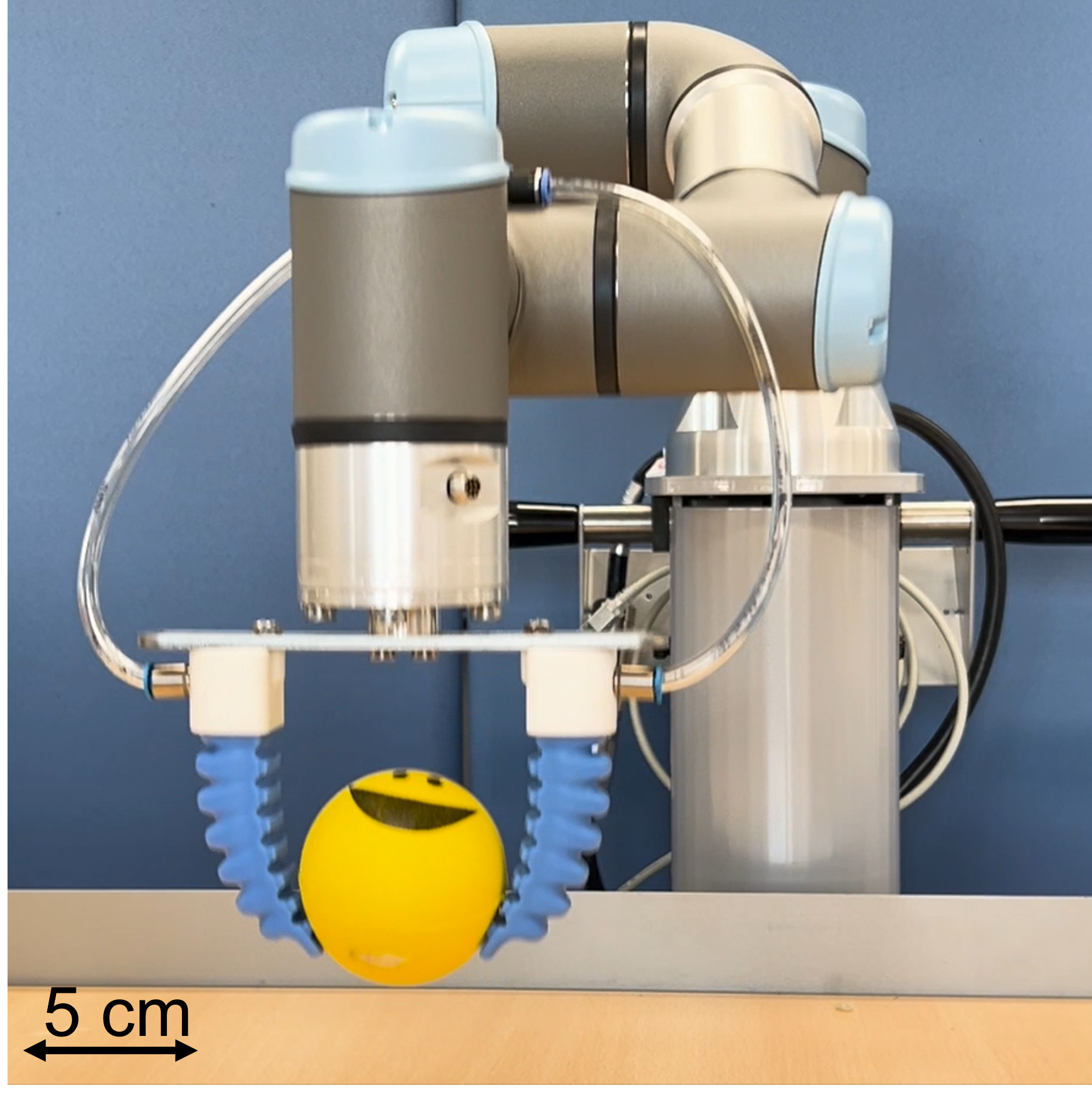}
    \caption{A pair of the real soft grippers mounted on a UR3e robotic arm. Pressurized air pipes are visible.}
    \label{fig:real_soft_gripper}
    \end{subfigure}
    \hfill
    \begin{subfigure}[t]{0.49\linewidth}
        \centering
        \includegraphics[width=\linewidth]{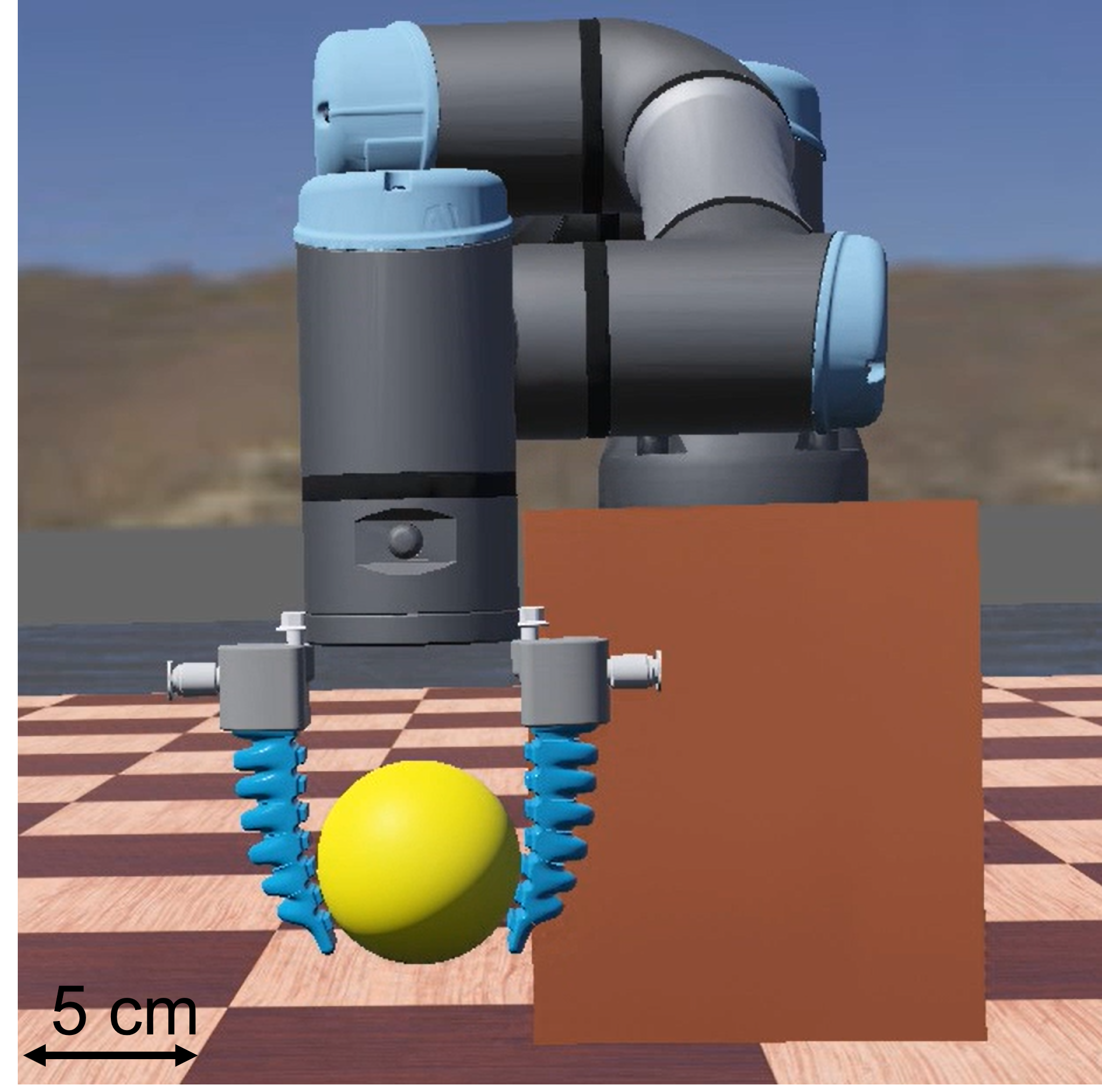}
    \caption{A pair of the simulated soft grippers mounted on a simulated UR3e robotic arm in Webots.}
    \label{fig:simulated_sof_gripper}
    \end{subfigure}
    \caption{Soft gripper attached to the UR3e robotic arm shown in the real setup and in the Webots simulation environment. The soft gripper is commercially available \cite{SoftGripping}.}
    \label{fig:Soft_gripper}
\end{figure}

Previous studies have explored standard solid mechanics discretization methods like Finite Element Analysis (FEA) \cite{Moseley} and voxel-based discretization \cite{Hu} to simulate soft robots. Other approaches, including the Cosserat rod theory \cite{Armanini} and the piece-wise constant curvature (PCC) approach \cite{Caasenbrood}, have also been studied to simulate soft robots' behavior. Data-driven approached also has been employed to model the kinematics of the soft gripper \cite{Rad}. However, these methods mainly serve as Computer-Aided Design (CAD) tools rather than comprehensive simulation environments for robots. Several dedicated simulation environments have been developed to simulate soft robots' behavior, such as SOFA \cite{Ferrentino}, primarily targeted at real-time physical simulation, and SoMo \cite{Graule}, a framework based on pyBullet. Mujoco \cite{Bednarek}, a general-purpose efficient physics engine designed to advance research and development in robotics, offers fluid dynamic simulation capabilities that are also usable for soft robot simulations. Other example of simulation environment include toolboxes within MATLAB that have been developed particularly for soft robotics simulation such as Sorotoki \cite{Caasenbrood2020}, which facilitate the development of soft robots by providing a comprehensive set of tools for design, modeling, and control, and also SoRoSim \cite{Anup}, that uses the Geometric Variable Strain (GVS) approach to provide a unified framework for the modeling, analysis, and control of soft robots. All of these simulation environments lack the comprehensiveness of full-fledged robotic simulators with integrated sensors, actuators, and environment modeling.

In this study, we focus on pneumatically actuated beam-like soft grippers, a well-established category of soft robots \cite{Galloway} \cite{Abondance}. We present a computationally lightweight physics-based simulation model for beam-like soft grippers, directly integrated within the Webots robotic simulator. This integration provides access to Webots's extensive sensor and actuator libraries. 
Furthermore, it offers capabilities of integrating soft and rigid robots, such as UR3e arm manipulator, within the same simulation environment, as shown in Figure \ref{fig:simulated_sof_gripper}. 

Firstly, we show how we model the structure of the soft body of a beam-like soft gripper via a Rigid-Link-Discretization (RLD) approach. The soft grippers are modeled as a series of rigid links, where each link is described by a mass-spring-damper system and is connected via joints to its neighboring links. Such a discretization method can readily be implemented in all open-source robotic simulation platforms. Secondly, we present an automated systems identification process based on dynamical data of the physical systems to create a digital twin instance of the physical soft grippers reducing the disparity between the simulation and reality. We use a heuristic search method, the Particles Swarm Optimization (PSO) algorithm to obtain a set of optimized parameters. Finally, we validate the accuracy of our simulation model using experimental data that are not used for systems identification. 

Our work offers a solution for harnessing the capabilities of the Webots robotic simulator in the study of beam-like soft robots that can be integrated or interact with rigid robots. This approach enables fully controllable and computationally efficient simulations of soft robots, coupled with an automatic calibration process to create comprehensive models integrated within an established robotic simulator.

\section{Modeling Approach}

To create lightweight simulation models with controllability features similar to those in robotic simulators for rigid-link robots, we develop our model directly within Webots. Webots is an open-source physics-based robotics simulator utilizing the Open Dynamics Engine (ODE) for simulating rigid body dynamics. Although Webots does not inherently support soft body dynamics, it includes a simulation environment parameter (CFM) that adjusts collision hardness, allowing rigid objects to pass through each other on impact. This provides a basic means to simulate the softness of contact points between solid objects.

\begin{figure}[t]
    \centering
    \begin{subfigure}[t]{0.49\linewidth}
        \centering
        \includegraphics[width=\linewidth]{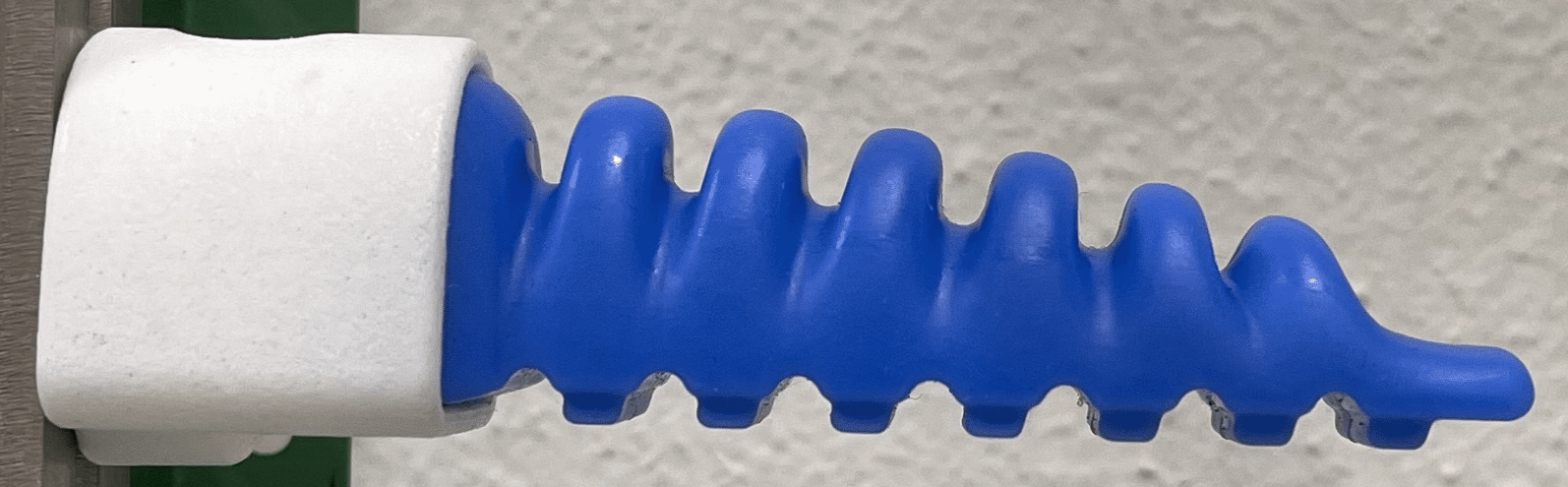}
        \caption{Real soft pneumatic gripper (SoftActuator Finger), the blue part is the soft flexible part.}
        \label{fig:Soft_gripper_real}
    \end{subfigure}
    \hfill
    \begin{subfigure}[t]{0.49\linewidth}
        \centering
        \includegraphics[width=\linewidth]{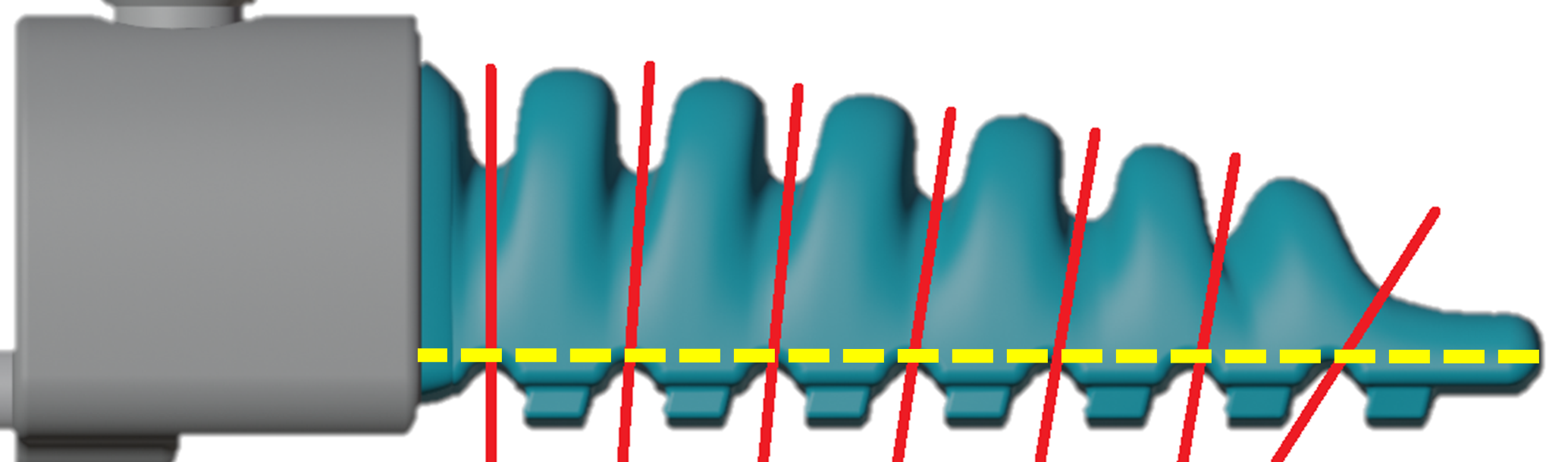}
        \caption{CAD model segment slicing, the red lines are the slicing lines and the yellow dashed line is the bending line where joints are.}
        \label{fig:CAD_slicing}
    \end{subfigure}
    \vskip\baselineskip
    \begin{subfigure}[t]{0.49\linewidth}
        \centering
        \includegraphics[width=\linewidth]{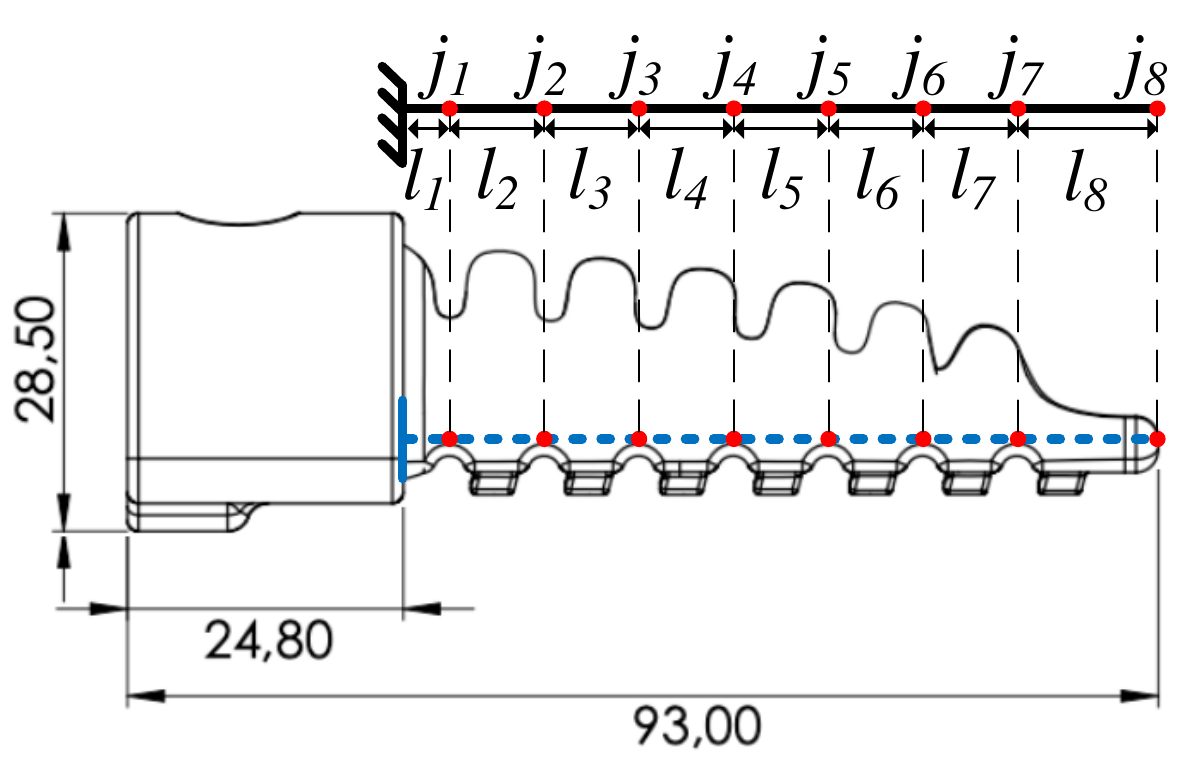}
        \caption{Soft gripper design side view with dimensions in mm, the blue dashed line is the bending line and red dots are joints positions.}
        \label{fig:diagram_cad}
    \end{subfigure}
    \hfill
    \begin{subfigure}[t]{0.49\linewidth}
        \centering
        \includegraphics[width=\linewidth]{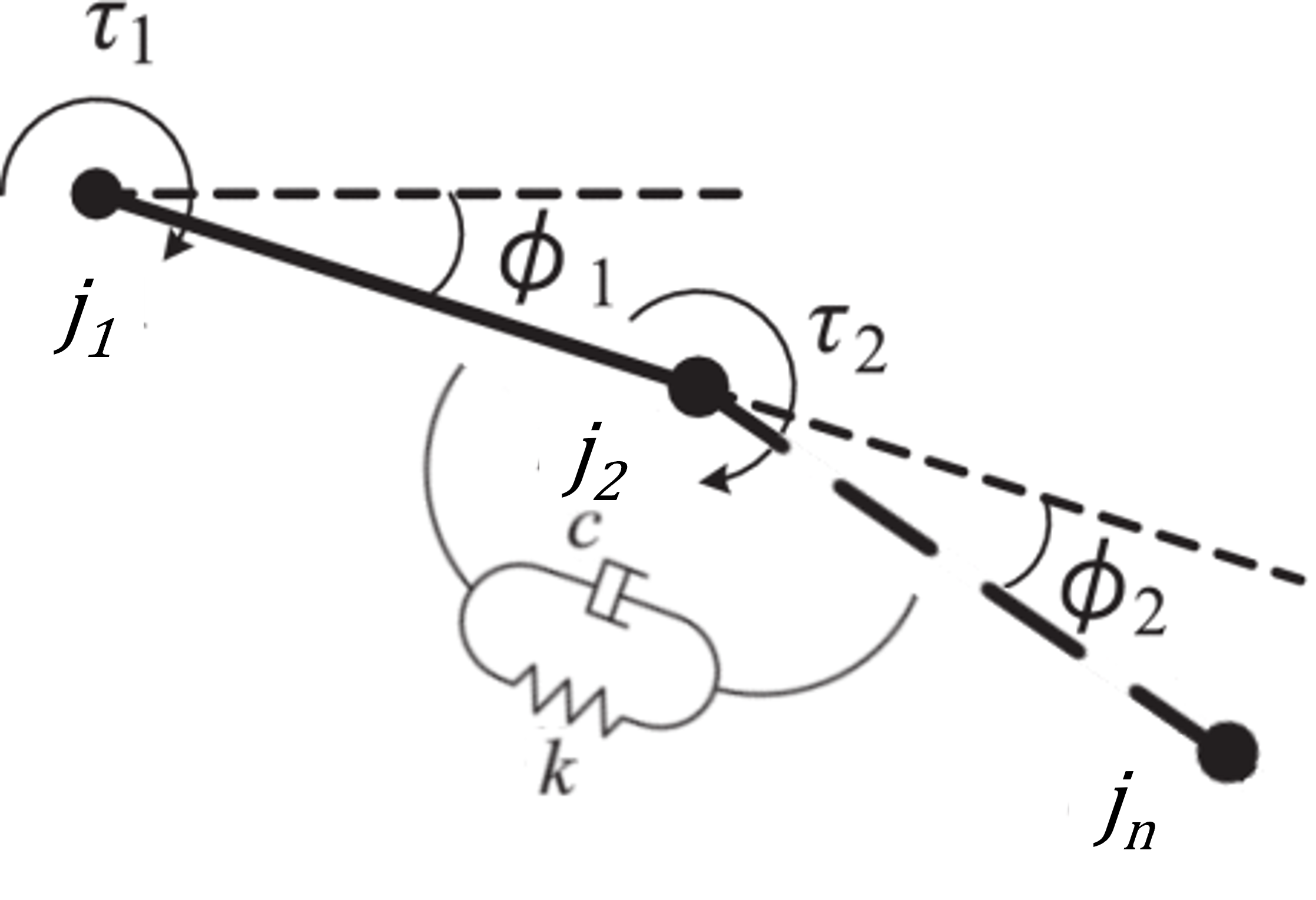}
        \caption{RLD model concept with virtual spring and damper on joints.}
        \label{fig:RLD_model}
    \end{subfigure}
    
    \caption{A depiction of (a) the real soft gripper, (b,c) CAD model slicing \cite{SoftGripping}, and (d) the RLD model concept.}
    \label{fig:soft_gripper_model}
\end{figure}

In this study, our reference is an off-the-shelf beam-like pneumatically actuated soft gripper (SoftGripping SoftActuator Finger \cite{SoftGripping}) measuring roughly 68mm × 20mm × 22mm and weighing 20g, made from silicone rubber material as seen in Figure \ref{fig:Soft_gripper_real}. We simulate a beam-like soft robot using an RLD model, similar to previous works on \cite{wang2017soft}, where segments are connected via hinge joints in Webots. First, we slice the soft gripper's CAD model into several segments and consider the joints placed at the slicing locations as in Figure \ref{fig:CAD_slicing}. We assume the soft gripper bends primarily at the valley points. Since our soft gripper has seven gaps, thus eight rigid segments are created in the simulation model. 

Webots has various types of virtual joints to simulate the connection between segments, such as single-axis hinge joints, dual-axis hinge joints, slider joints, and ball joints. In our simulation, each segment is connected with a hinge joint, allowing rotation around the axis perpendicular to the side plane of the segments. The hinge joint node in Webots allows rotation along a main axis and is defined by two parameters: (i) the joint stiffness or spring constant (N.m/rad), and (ii) the joint friction or damping constant (N.m.s/rad). The spring constant controls the joint's springiness, emulating a torsional spring. A higher spring constant means a stiffer joint that resists bending. The damping constant controls energy dissipation in the joint, applying a force proportional to the joint's velocity: $F = -Bv$, where $B$ is the damping constant and $v = \frac{\dd x}{\dd t}$ is the joint velocity calculated by the physics simulator. The hinge joint node also includes a rotational motor that can be actuated via torque input to control the joint angle. When it is actuated, the joint angle follows the motor input (active joint); otherwise (when not actuated), it responds to the net torque (passive joint) \cite{cyberbotics_rotationalmotor}. 


The mass-spring-damper system in Webots simulates a beam-like soft robot by linking rigid segments via hinge joints as shown in Figure \ref{fig:RLD_model}. These joints enable rotation around an axis and are characterized by joint stiffness (spring constant) and joint friction (damping constant) within Webots. 
We used the default softCFM value of 0.001. We add extra weight to the tip of the soft gripper to create a bending curve to measure the spring constant. We also bend the soft gripper to the initial position before suddenly releasing the soft gripper to measure the damping constant. 
For a given damping constant, the soft gripper's static behavior is influenced by the spring constant; a higher spring constant makes the beam stiffer and less droopy. For a given spring constant, the soft gripper's dynamic behavior is affected by the damping constant; a higher damping constant causes the beam to reach a steady state after a perturbation slowly with fewer oscillations. For the active behavior while the gripper is under pneumatic actuation, simulated by the rotational motor with torque constant in each joint within Webots. The torque constant determines the torque that applies in each joint regarding the real pressure that is applied on the gripper. Section III details how we calibrate these parameters by comparing the static and dynamic behaviors of the simulated beam against those of a real soft beam.


\section{Calibration Approach}
The calibration process 
involves finding model parameters to match measurements between real-world experiments and simulated ones. We consider passive (no pneumatic actuation) experiments for calibrating the spring and damping parameters and active (with pneumatic actuation) experiments for calibrating the torque parameters. We use simulations in Webots to compare and iteratively adjust the model parameters to align real and simulated behavior. This process involves using a Particle Swarm Optimization (PSO) method for optimization of the parameters and simulations in Webots, where each of the particles in the PSO swarm corresponds to a Webots simulation world.

\subsection{Automatic Calibration Procedure} 
The primary objective of our calibration routine is to identify parameters that allow for the closest match of measurements between real-world experiments and simulated experiments in Webots. We use a PSO heuristic search algorithm to efficiently cover the search space for each experiment. PSO is inspired by the social behavior of birds and fish, and finds locally optimal solutions through collaboration among particles in a multidimensional search space \cite{Poli}. In this approach, each individual within the population, referred to as a particle, represents a distinct instance of our simulated world characterized by a specific set of parameters for the RLD model. Each particle's position $x_i$ and velocity are initialized randomly on an appropriate range. 
Thereafter, in each iteration, the fitness of each particle is assessed through a measure that is able to capture the similarity between real-world experiments and simulated ones. Subsequently,  in each iteration, each particle's position and velocity are updated:

\begingroup
\small
\begin{align}
\mathbf{v}_i(t+1) &= \omega \mathbf{v}_i(t) + c_1 r_1 (\mathbf{p}_i - \mathbf{x}_i(t)) + c_2 r_2 (\mathbf{g} - \mathbf{x}_i(t))
\label{velocity update} \\
\mathbf{x}_i(t+1) &= \mathbf{x}_i(t) + \mathbf{v}_i(t+1)
\label{position update}
\end{align}
\endgroup
where $ \omega $ is the inertia weight,  $c_1 $ and $ c_2 $ are cognitive and social coefficients, $ r_1 $ and $ r_2 $ are random numbers uniformly distributed in [0, 1], $ \mathbf{p_i} $ is a particle's personal best and $ \mathbf{g} $ is the global best position. We use an implementation of PSO found in the pyswarms library \cite{Miranda}, outlined in Algorithm \ref{alg:PSO_alg}.

The calibration routine is automated using Python scripts. The parameters for the calibration are loaded from a JSON file (parameters.txt). This file contains critical information such as the type of calibration (spring or damping), and optimizer settings. 
To enhance the efficiency of the calibration process, multi-threading is employed. This allows multiple simulations to run concurrently, leveraging the computational power of modern multi-core processors. Threads are controlled with a locking mechanism to ensure that fitness calculations are synchronized and that each particle's fitness score is accurately updated. 

\begin{algorithm}[t]
\caption{Particle Swarm Optimization (PSO)}
\begin{algorithmic}

\State \textbf{Initialization:}
\For{each particle $i$}
    \State Initialize position vector $\mathbf{x}_i$ and velocity vector $\mathbf{v}_i$
    \State Set best-known position $\mathbf{p}_i$ to initial position $\mathbf{x}_i$
\EndFor
\State \textbf{PSO iterations:}
\While{termination criterion not met}
    \For{each particle $i$}
        \State Update velocity
        \State Update position
    \EndFor
    \For{each particle $i$}
        \State Evaluate fitness using the objective function
    \EndFor
\EndWhile

\end{algorithmic}
\label{alg:PSO_alg}
\end{algorithm}

\subsection{Passive Calibration Experiments} 

\begin{figure}[t]
    \centering
    \begin{subfigure}[b]{0.49\linewidth}
        \centering
        \includegraphics[width=\linewidth]{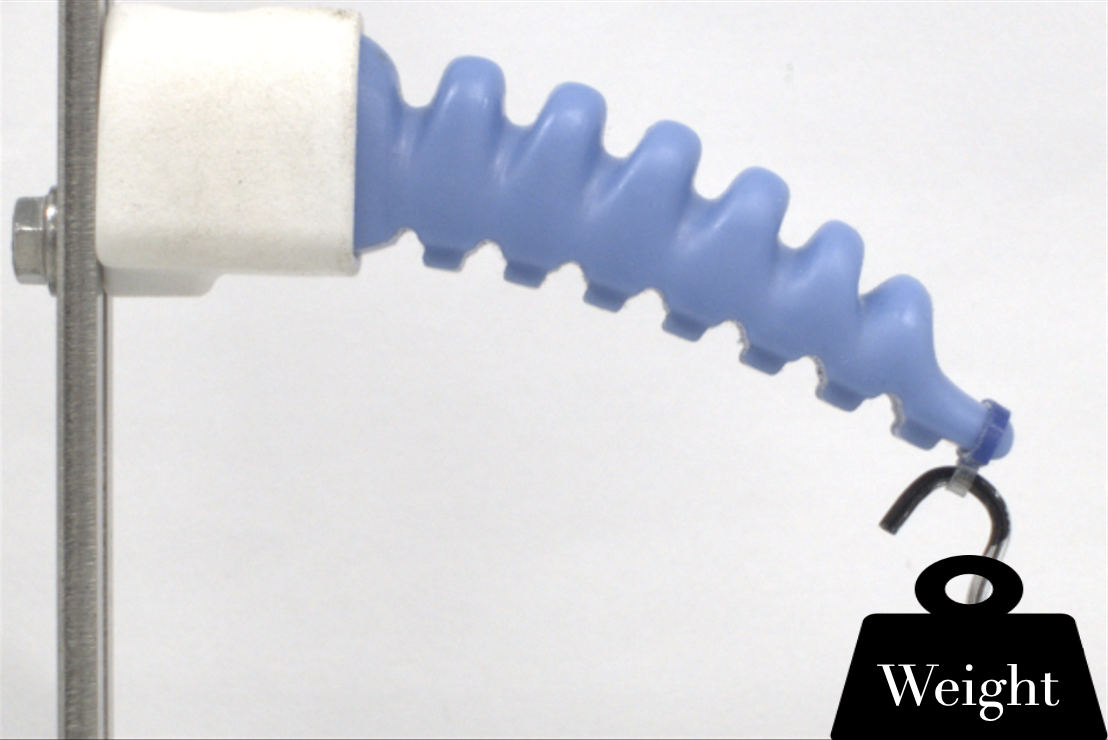}
        \caption{Original image with loading.}
        \label{fig:im_pr_1}
    \end{subfigure}
    \hfill
    \begin{subfigure}[b]{0.49\linewidth}
        \centering
        \includegraphics[width=\linewidth]{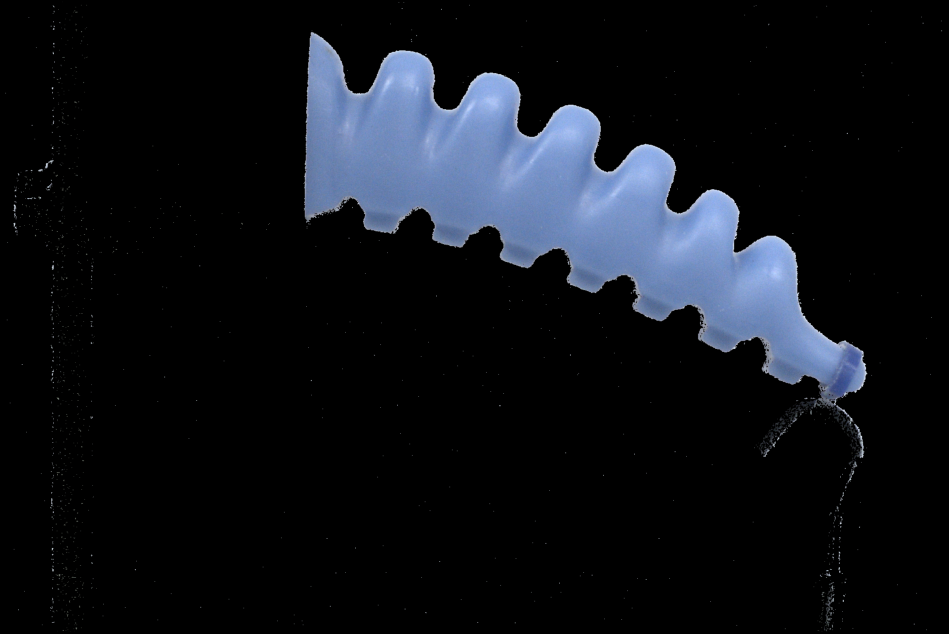}
        \caption{Combined mask on image.}
        \label{fig:im_pr_4}
    \end{subfigure}
    \begin{subfigure}[b]{0.49\linewidth}
        \centering
        \includegraphics[width=\linewidth]{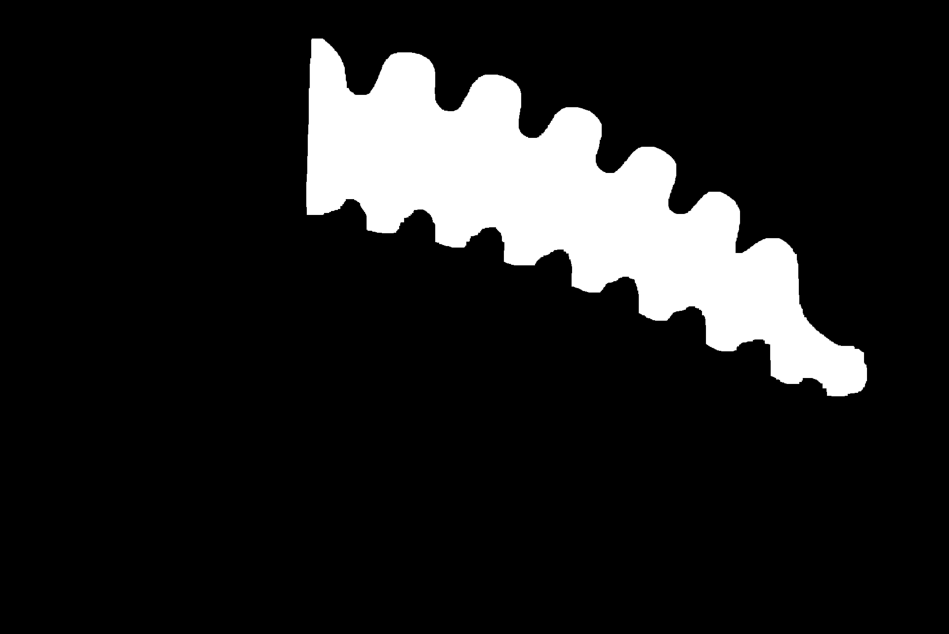}
        \caption{Binary threshold on mask.}
        \label{fig:im_pr_5}
    \end{subfigure}
    \hfill
    \begin{subfigure}[b]{0.49\linewidth}
        \centering
        \includegraphics[width=\linewidth]{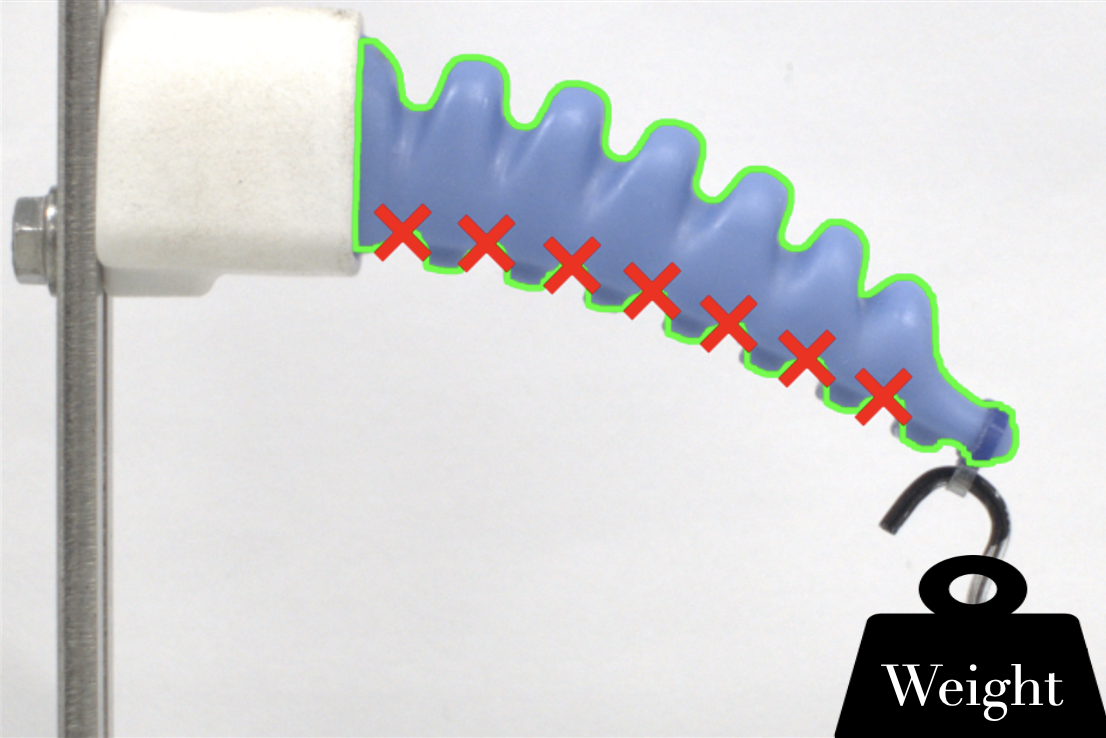}
        \caption{Original image with annotated contour and joint locations}
        \label{fig:im_pr_6}
    \end{subfigure}
    \caption{Image processing steps to assign the joint locations.}
    \label{fig:image_proc}
\end{figure}

Passive calibration determines model parameters by subjecting the gripper to an external load without pneumatic actuation to analyze its static and dynamic response. We conduct static and dynamic passive experiments with the gripper positioned downward as seen in Figure \ref{fig:image_proc}. 

In a first step, we calibrate the spring constant of the soft gripper in a static experiment by subjecting the gripper to an external load at the tip. An image is then taken using a camera setup at a fixed distance and angle with respect to the soft gripper. The images are processed to obtain the rotational angle of each joint. This experiment is repeated with various weights added to the tip of the gripper. A virtual position sensor is used to measure a joint's angle within Webots. 

The image processing workflow begins by converting the original image to grayscale and applying a threshold to isolate non-white elements. The original image is then transformed into the HSV color space, and a mask is applied to retain only the gripper’s hue, saturation, and value within a specified range. Advantages of using HSV are that it presents immunity to illumination changes and the color hue changes are continuous and linear \cite{chavolla2018color}. This HSV mask is combined with the grayscale threshold mask and applied to the original image. The resulting image is further processed by converting it to a binary grayscale format to clearly outline the gripper’s shape. Additional morphological operations like dilation and erosion can be performed to reduce noise and better define object boundaries \cite{morph} \cite{jackway1996scale}. In the binary image, the contour of the gripper, defined as the curve that joins all continuous points with the same color or intensity, is detected using OpenCV's findContours function \cite{opencv_contours_tutorial}. The contour is split to obtain the bottom- and top line of the contour by selecting the minimal and maximal x-value. Finally, the peaks of the bottom contour, representing joint locations, are identified  through comparison of neighboring values \cite{scipy_find_peaks_tutorial}, and the angles between these joints are calculated following:

\begin{equation}
    \theta_i =
    \begin{cases}
        \arctan\left(\frac{-\Delta y_i}{\Delta x_i}\right) & \text{if } i = 1 \\
        \arctan\left(\frac{-\Delta y_i}{\Delta x_i}\right) - \sum_{j=1}^{i-1} \theta_j & \text{if } i > 1
\end{cases}
    \label{inv_kin}
\end{equation}
where $(x,y)$ represents the location of a joint and $\theta_i\in [-\frac{\pi}{2},\frac{\pi}{2}]$ is the $i^{th}$ joint angle. These steps are shown in Figure \ref{fig:image_proc}.

The fitness of a parameter set is computed as the mean squared error of the difference in joints' simulated and real angular position:
\begin{equation}
    \text{Fitness} = \frac{1}{N} \sum_{j=1}^{N} (\phi_{j}^{\text{sim}} - \phi_{j}^{\text{real}})^2
    \label{eq:fitness_spring}
\end{equation}
where $ N $ is the number of joints, $ \phi_{j}^{\text{sim}} $ is the angle of joint $ j $ in the simulation, and $ \phi_{j}^{\text{real}} $ is the angle of joint $ j $ in the real experiment. Lower values denote a closer matching.

In the second step, we calibrate the damping constant of the soft gripper's joints in the dynamic experiment. The gripper is subjected to a step load and we compare the gripper's settling time and number of overshoots. In the real experiment, we apply a load on the gripper's tip as an initial condition and then instantly remove the load to observe the settling behavior of the gripper. In the simulated experiment, we replicate similar conditions using a virtual load applied at the tip. Webots does not allow for the releasing of an object, instead, we first obtain the joint angles in an initial simulation and use these as an initial condition in subsequent experiments to observe the settling behavior of the gripper.

\begin{figure}[t]
    \centering
    \includegraphics[width=0.8\linewidth]{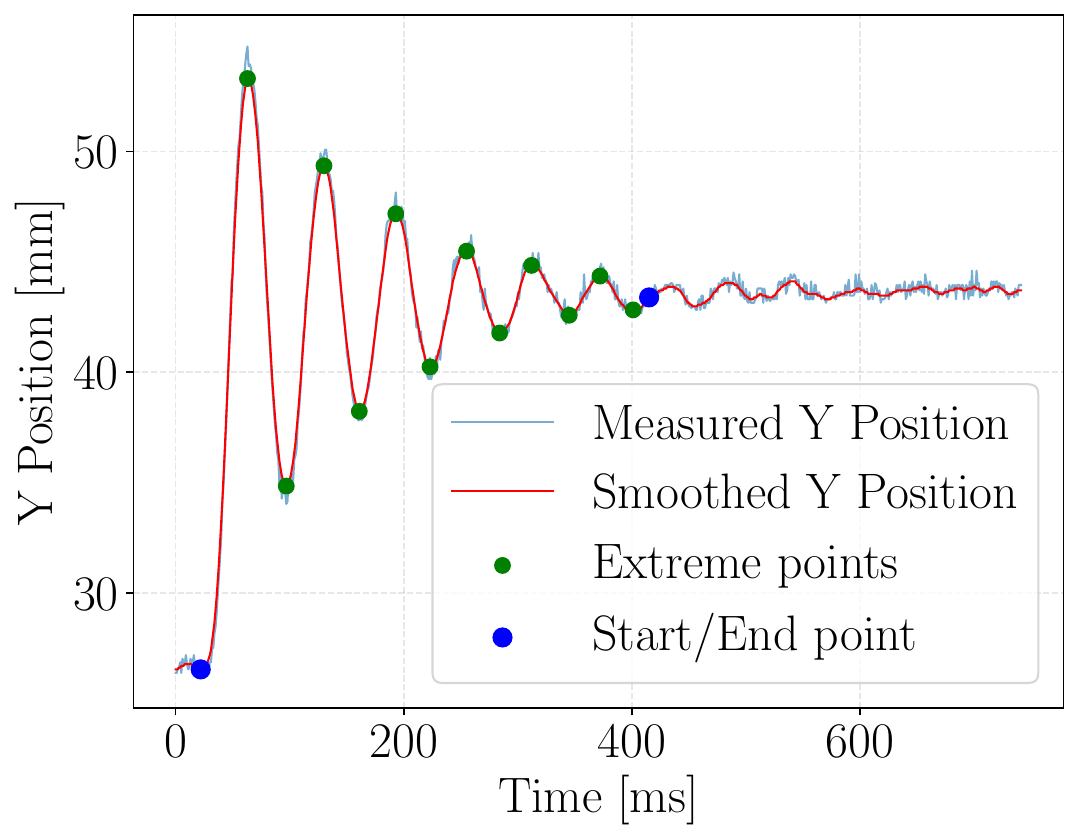}
    \caption{Settling behavior of the gripper after instantaneous release of an added weight at the tip: Before release, the gripper was subjected to 40 grams of added weight to the tip. The number of overshoots in this experiment was 12 and the settling time 0.409 seconds.}
    \label{fig:video_processing}
\end{figure}

The gripper's response is measured by tracking the vertical position of the corner with the highest x-value in a video recording. To capture the grippers behavior, the Sony ZV-1 camera was used \cite{sony-zv1-website}. An effective frame rate of 960-1000 frames per second was employed. In real experiments, this is done using an automated video processing function that detects the gripper’s contour in each frame with the aforementioned techniques. In simulated experiments, the location is obtained by using a forward kinematics scheme from the rotational angles in the joints and the length between them. The detected y-positions of the real experiment are smoothed with a Gaussian filter \cite{scipy_gaussian_filter}, and key extrema (peaks and valleys) are identified to count overshoots and measure settling time \cite{scipy_find_peaks_tutorial}. We count an overshoot if the vertical distance between two overshoots is more then 1 millimeter. An example of the of the overshoot counting is plotted in Figure \ref{fig:video_processing}.

The fitness between the real-world and simulated experiments is computed as the following:
\begin{equation}
    \text{Fitness} =|T_{s}^{\text{sim}}-T_{s}^{\text{real}}|(1+|n_o^{\text{sim}}-n_o^{\text{real}}|)
    \label{eq:fitness_damping}
\end{equation}
where $T_s^{\text{sim}}$ and $T_s^{\text{real}}$ represent the simulated and real settling times and $n_o^{sim}$ and $n_o^{real}$ represent the number of overshoots in simulation and reality, respectively.

\subsection{Active Calibration Experiments} 
In active calibration experiments, we study the behavior of the gripper under pneumatic actuation to find torque parameters. The gripper is operated through pressurized air, where increasing pressure leads to flexing, and decreasing it results in the extension of the gripper. 
Active calibration involves applying forces and gauging the response to validate the soft gripper’s torque parameters. We applied pressurized air as an input for the real experiment between 0-100 KPa. We use virtual rotational motor torque to actuate the gripper in the simulator. The experimental setup is shown in Figure \ref{fig:active_exp}. The input variables for calibration encompass applied force, 
spring constant, 
and damping
. Output variables determining the fitness function for optimization involve force at the tip 
and deformation
. We used a scale to measure the force on the tip in the real experiment and a virtual force sensor in the simulation. As shown in Figure \ref{fig:active_exp}, the relationship between the applied pressure and the force exerted at the tip is approximately linear, consistent with previous research \cite{wang2017soft}. Thus, we model the joint torque as: 
\begin{equation} \tau_i = \alpha_i P \end{equation} 
where $\tau_i\ [Nm]$ is the torque applied at joint $i$, $\alpha_i\ [\frac{1}{m^2}]$ is the parameter to be calibrated, and $P\ [Pa]$ is the applied pressure. The fitness score is calculated as the difference between the force applied at the tip in real and simulation experiments:

\begin{figure}[t]
   \centering
    \begin{subfigure}[t]{0.41\linewidth}
        \centering
    \includegraphics[width=\linewidth]{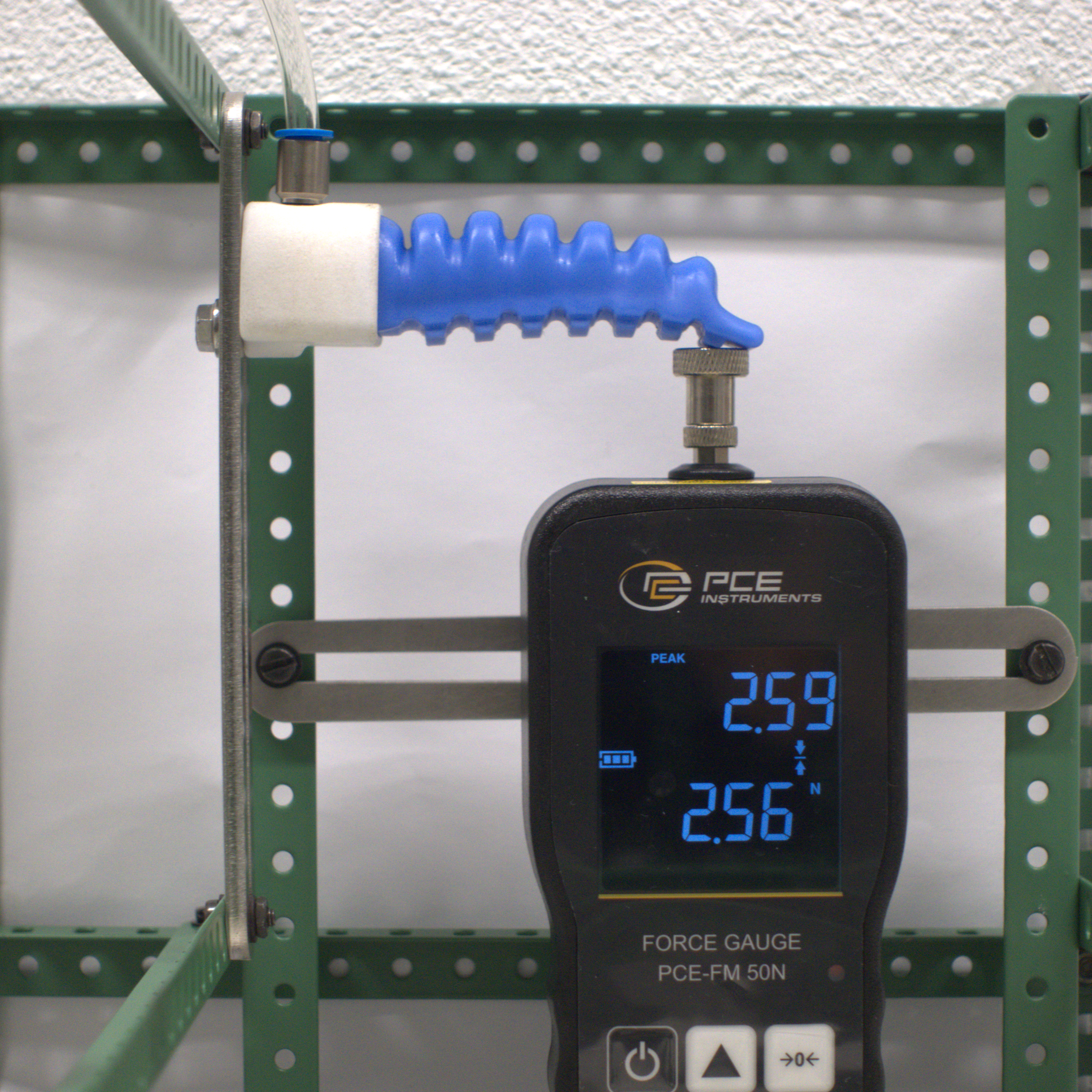}
    \caption{Force measurement at the gripper's tip with 100 KPa input air pressure.}
    \label{fig:force_gauge}
    \end{subfigure}
    \hfill
    \begin{subfigure}[t]{0.57\linewidth}
        \centering
    \includegraphics[width=\linewidth]{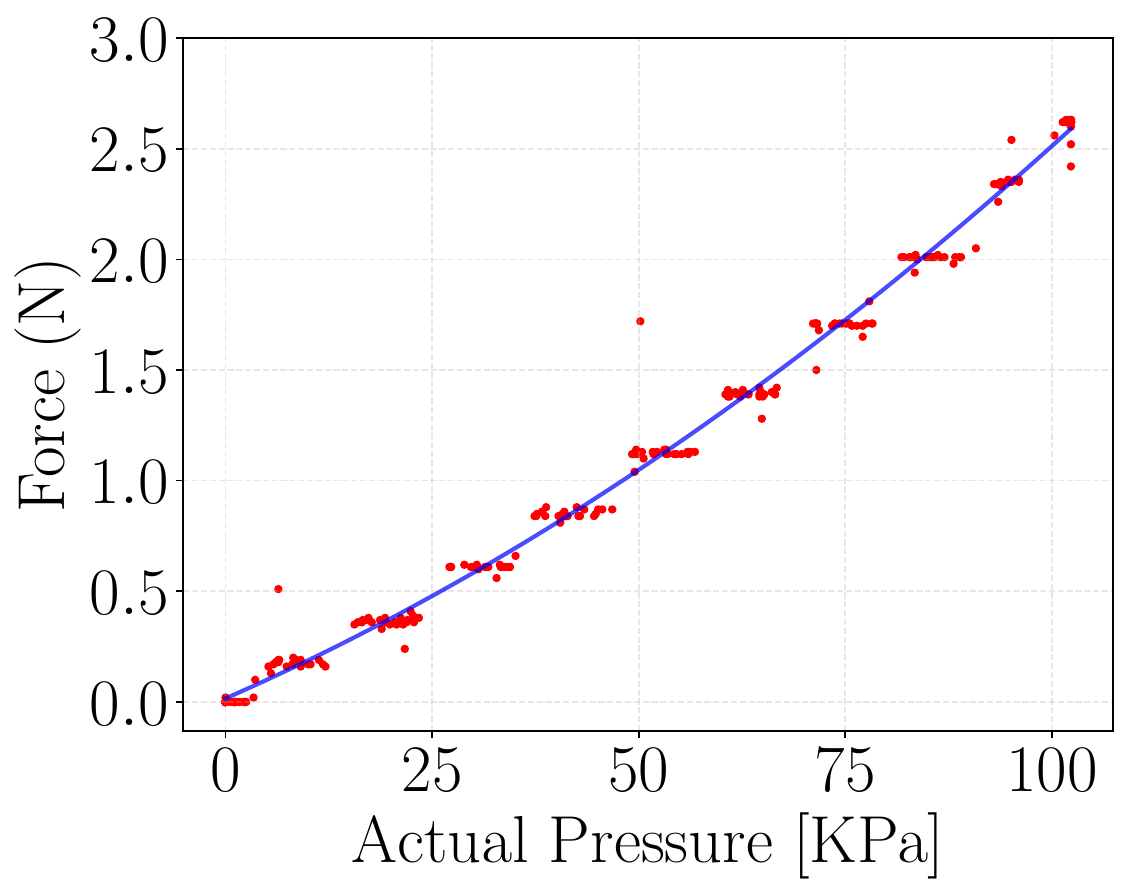}
    \caption{Relationship between the applied pressure and the exerted force at the tip of the gripper}
    \label{fig:real_active}.
    \end{subfigure}
    \caption{Active experiment force measurements.}
    \label{fig:active_exp}
\end{figure}

\begin{equation}
    \text{Fitness} = |F^{\text{sim}}-F^{\text{real}}|
    \label{eq:fitness_torque}
\end{equation}
where $F^{\text{sim}}$ and $F^{\text{real}}$ are the force that is exerted by the virtual gripper and real gripper respectively. 

\section{EXPERIMENTAL SETUP AND RESULTS}

\begin{figure}[t]
    \centering
    \begin{subfigure}[b]{0.49\linewidth}
        \centering
        \includegraphics[width=\linewidth]{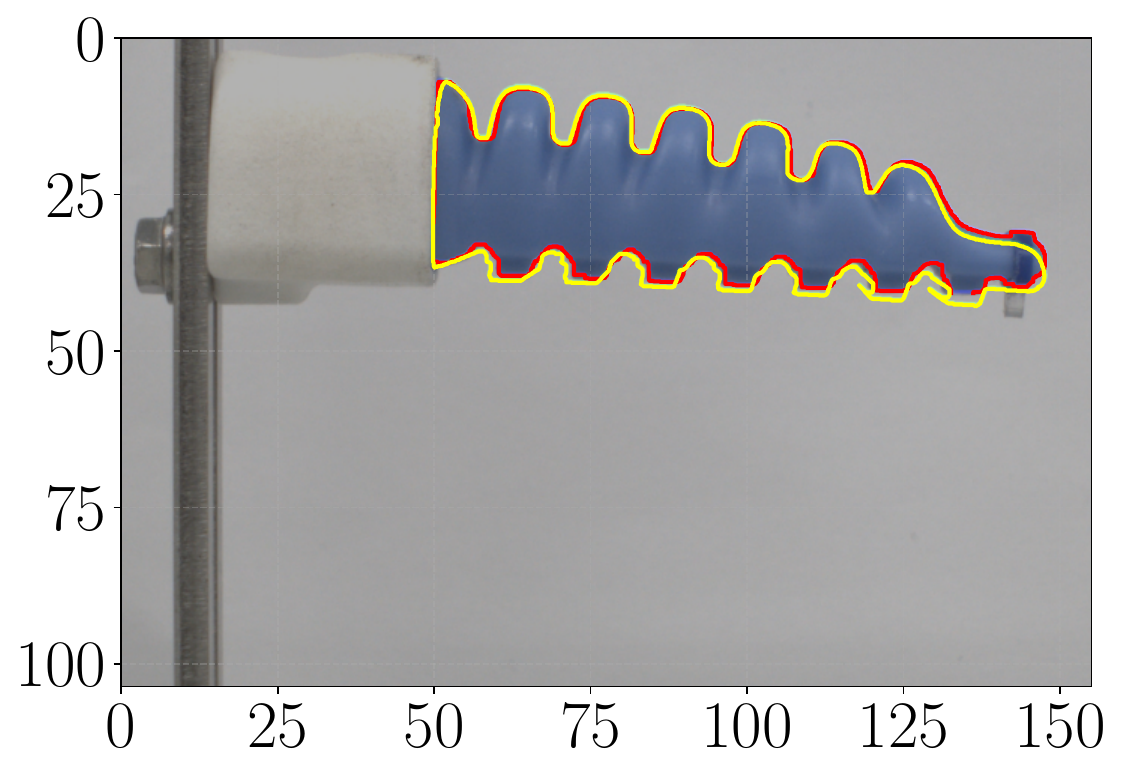}
        \caption{0 grams additional mass.}
        \label{fig:overlay_0}
    \end{subfigure}
    \hfill
    \begin{subfigure}[b]{0.49\linewidth}
        \centering
        \includegraphics[width=\linewidth]{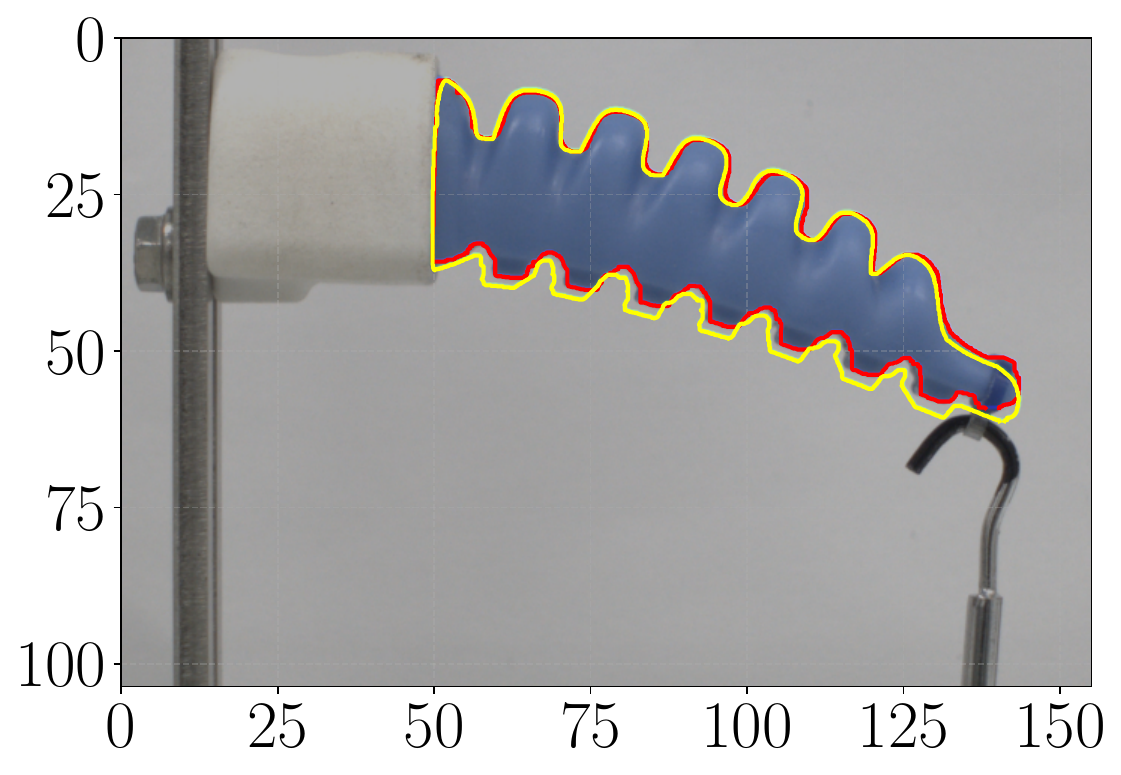}
        \caption{30 grams additional mass.}
        \label{fig:overlay_30}
    \end{subfigure}
    \begin{subfigure}[b]{0.49\linewidth}
        \centering
        \includegraphics[width=\linewidth]{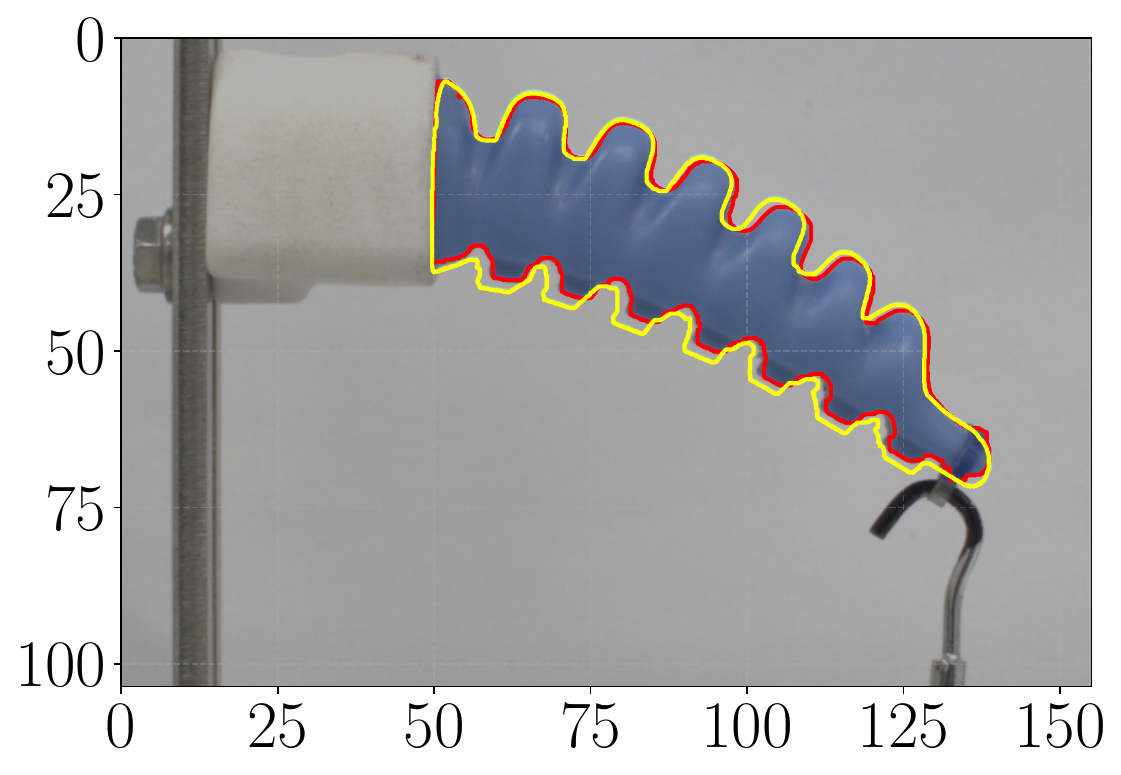}
        \caption{50 grams additional mass.}
        \label{fig:overlay_50}
    \end{subfigure}
    \hfill
    \begin{subfigure}[b]{0.49\linewidth}
        \centering
        \includegraphics[width=\linewidth]{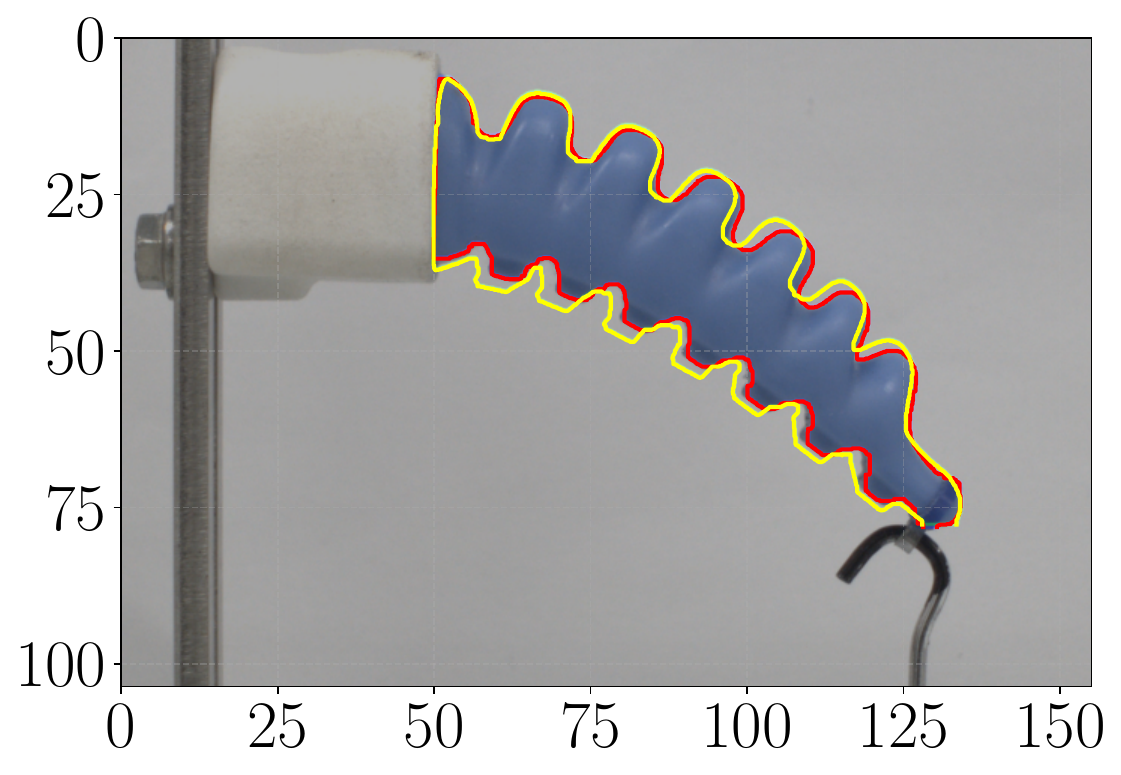}
        \caption{70 grams additional mass.}
        \label{fig:overlay_70}
    \end{subfigure}
    \caption{Validation of calibration routine for passive static experiments: Soft gripper subjected to a load in the real-world experiment with a contour overlay of its simulated counterpart with identical load. The red line indicates the contour obtained from the real-world image, while the yellow line represents the contour derived from the Webots simulation analysis. Both axes are measured in millimeters.}
    \label{fig:passive_exp_val}
\end{figure}

\begin{figure}[!ht]
    \centering
    \begin{subfigure}[b]{0.49\linewidth}
        \centering
        \includegraphics[height=0.75\linewidth]{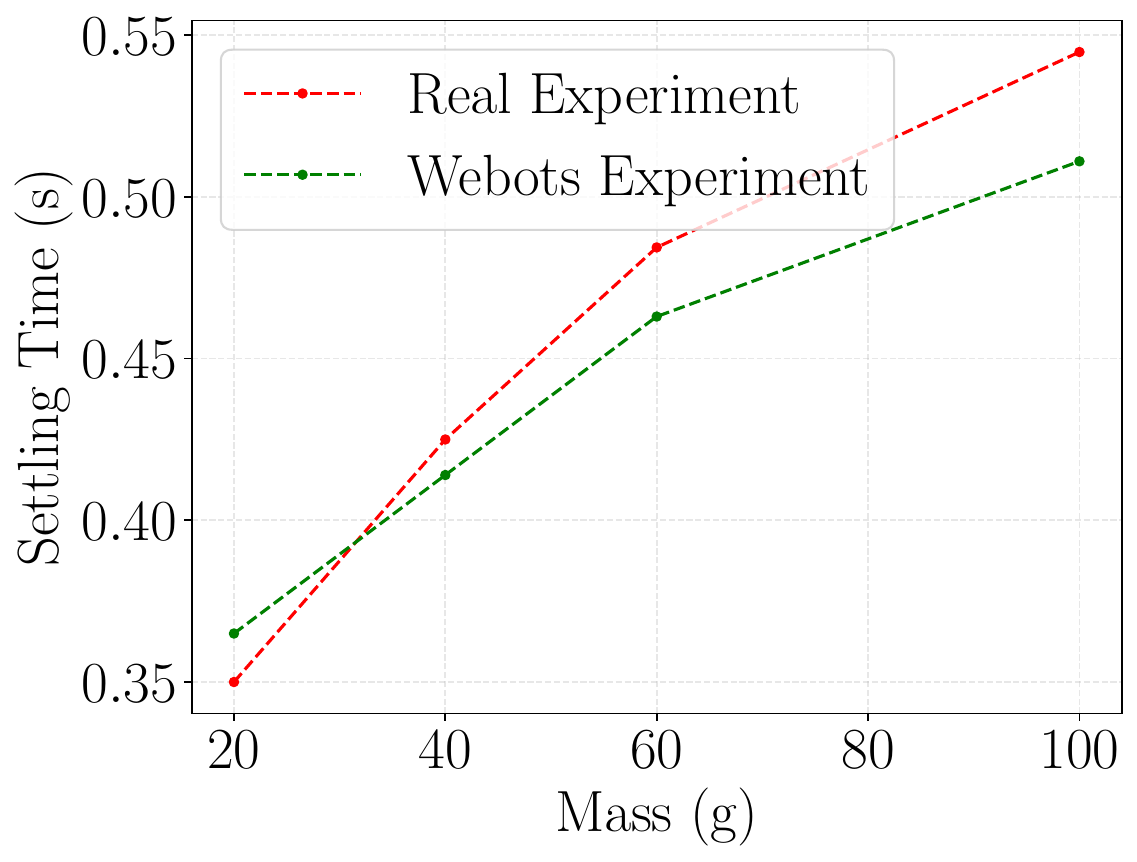}
        \caption{Settling time comparison.}
        \label{fig:settling_time_comparison}
    \end{subfigure}
    \hfill
    \begin{subfigure}[b]{0.49\linewidth}
        \centering
        \includegraphics[height=0.75\linewidth]{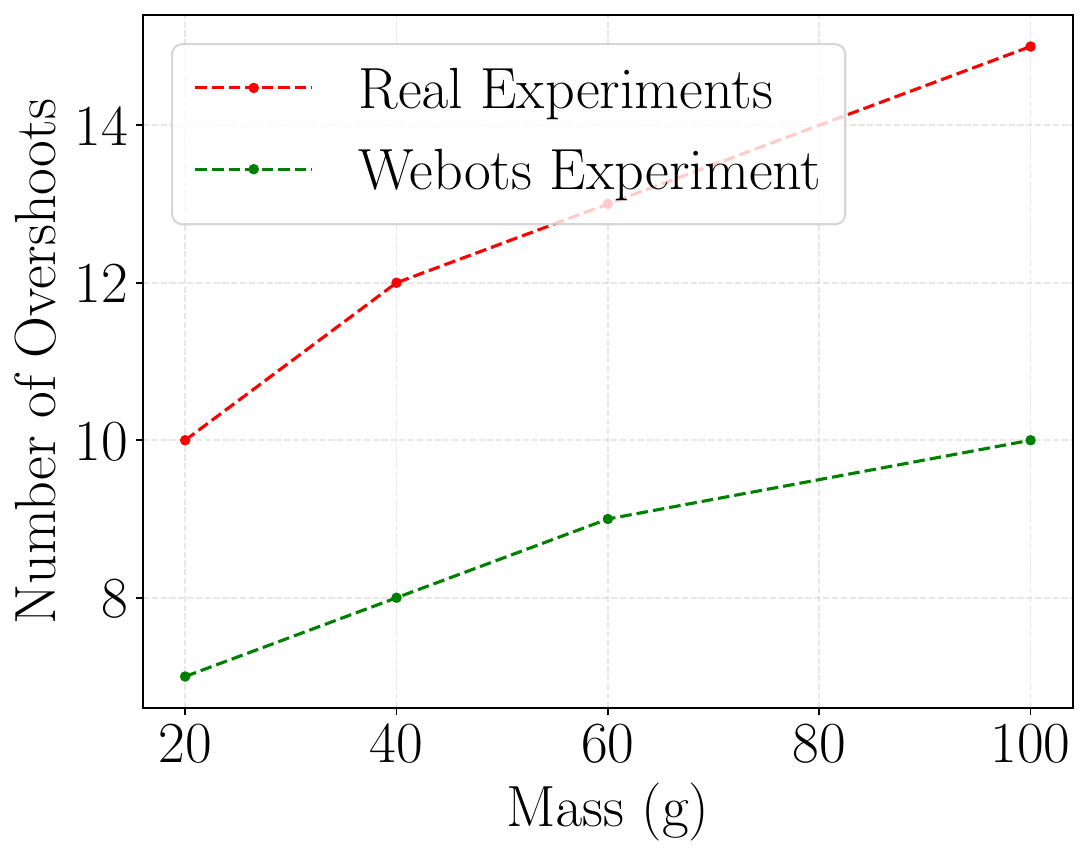}
        \caption{Overshoots comparison.}
        \label{fig:overshoots_comparison}
    \end{subfigure}
    \label{fig:passive_comparison}

     \begin{subfigure}[b]{0.49\linewidth}
        \includegraphics[height=0.75\linewidth]{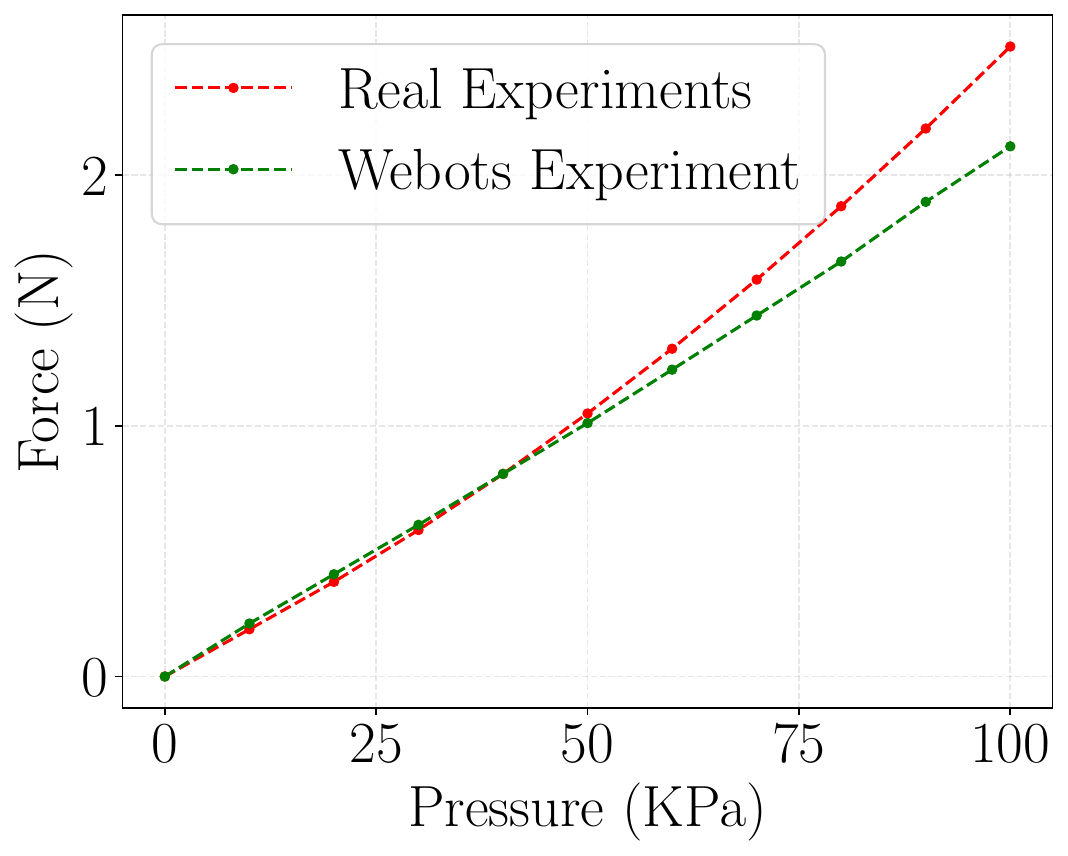}
        \caption{Force-pressure comparison.}
        \label{fig:active_comp}
     \end{subfigure}
     \caption{Validation of calibration routines for passive dynamic (a,b) and active experiments (c). For the passive dynamic experiment validation, settling times and the number of overshoots were analyzed under different externally applied loads. In the active experiment validation, the exerted force at the gripper's tip was compared across various applied pressures and corresponding torque coefficients in Webots.}

\end{figure}


\begin{figure*}[t]
    \centering
    \begin{subfigure}[b]{0.32\textwidth}
        \centering
        \includegraphics[width=\linewidth]{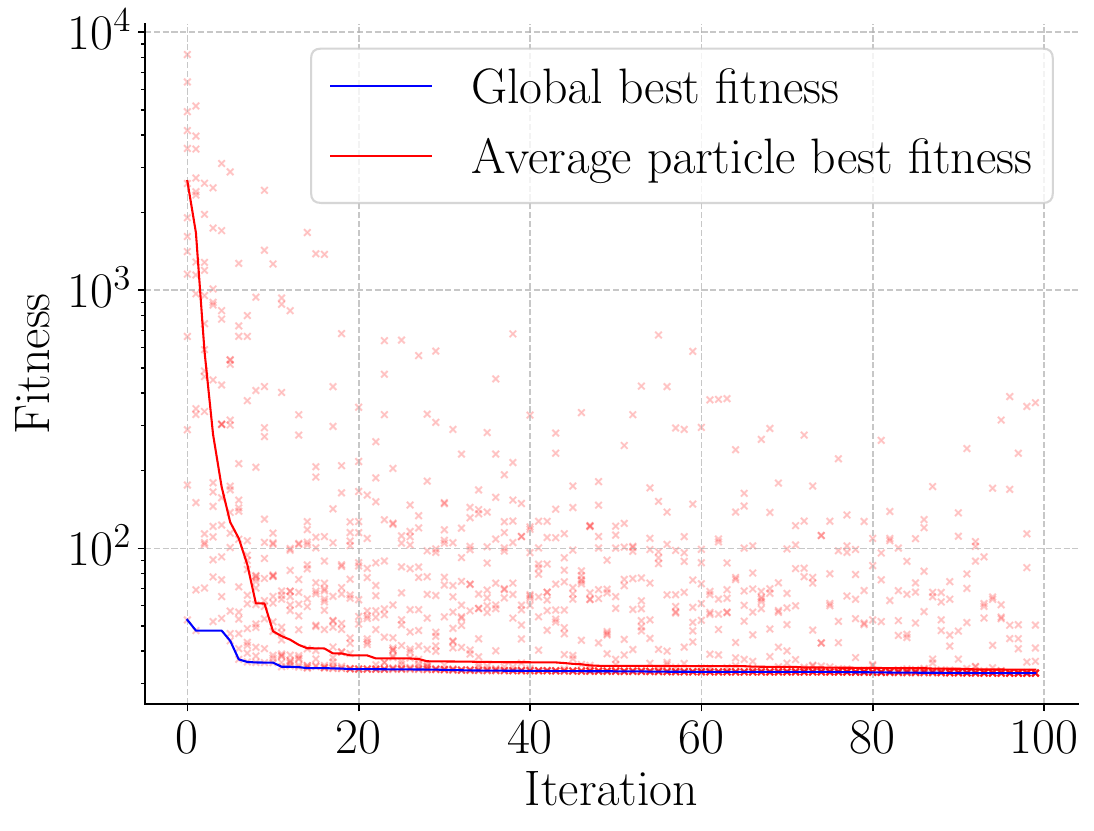}
        \caption{PSO process for spring parameters.}
        \label{fig:PSO_result_spring_downward}
    \end{subfigure}
    \begin{subfigure}[b]{0.32\textwidth}
        \centering
        \includegraphics[width=\linewidth]{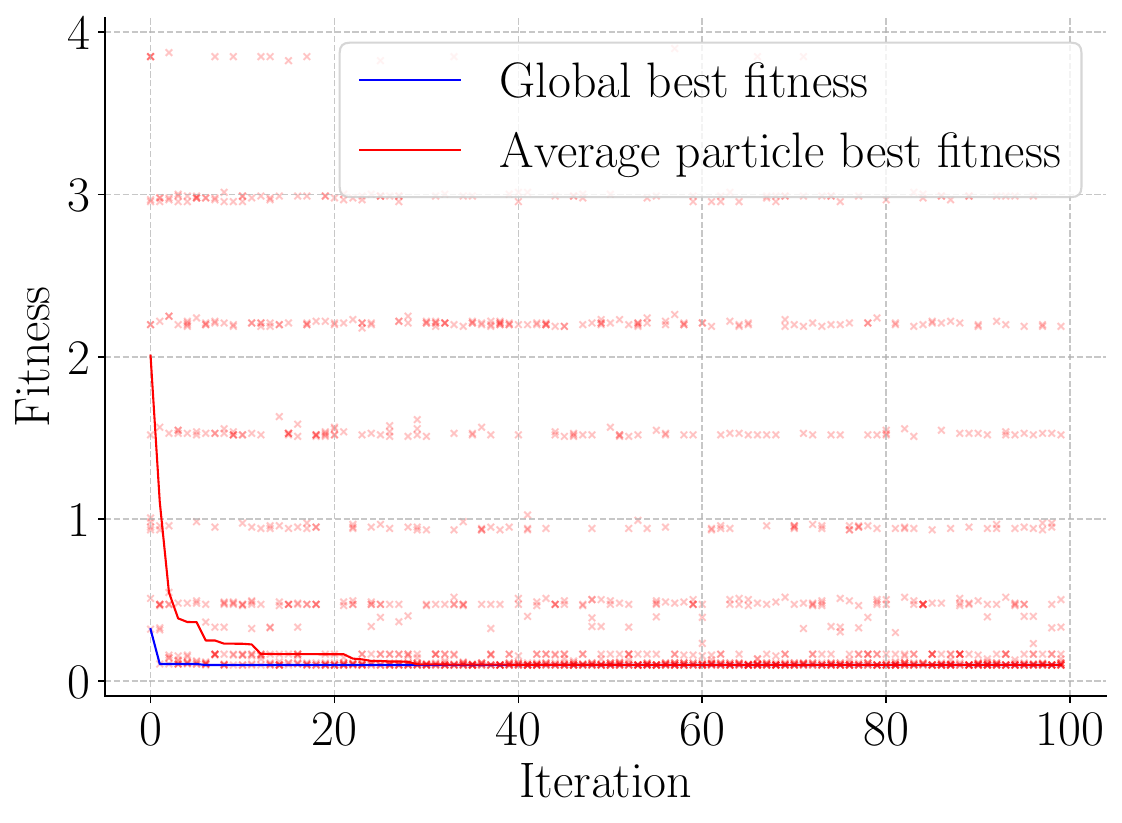}
        \caption{PSO process for damping parameters.}
        \label{fig:PSO_result_damping_downward}
    \end{subfigure}
    \begin{subfigure}[b]{0.32\textwidth}
        \centering
        \includegraphics[width=\linewidth]{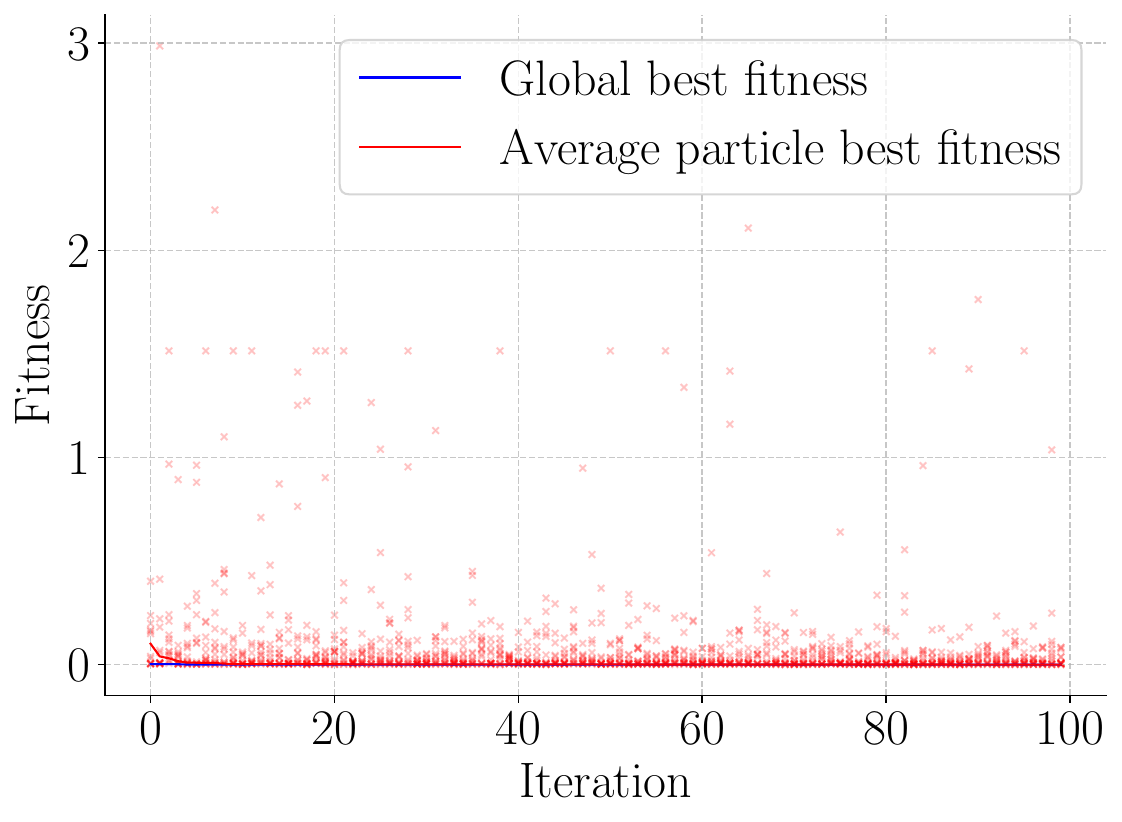}
        \caption{PSO process for torque parameters.}
        \label{fig:PSO_result_torque}
    \end{subfigure}
    \caption{We perform calibration using a PSO method to obtain model parameters of spring, damping, and torque. The line and scatter plot show the fitness evolution of particles during the PSO process.}
    \label{fig:PSO_results_downward}
\end{figure*}

\begin{figure*}[t]
    \centering
    \begin{subfigure}[b]{0.32\textwidth}
        \centering
        \includegraphics[height=5.5cm]{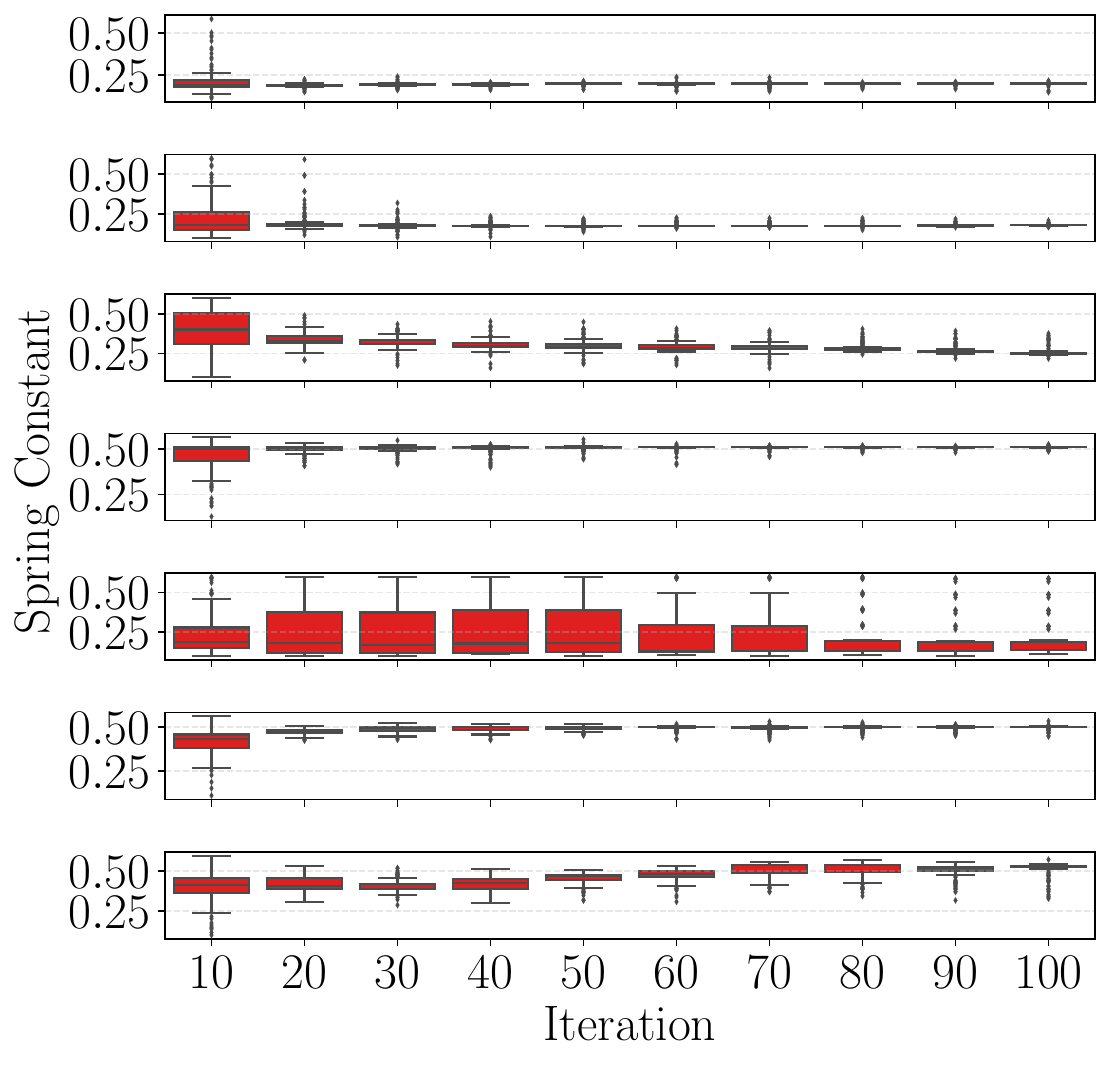}
        \caption{Spring constant convergence}
        \label{fig:PSO_result_spring_downward_boxplot}
    \end{subfigure}
    \begin{subfigure}[b]{0.32\textwidth}
        \centering
        \includegraphics[height=5.5cm]{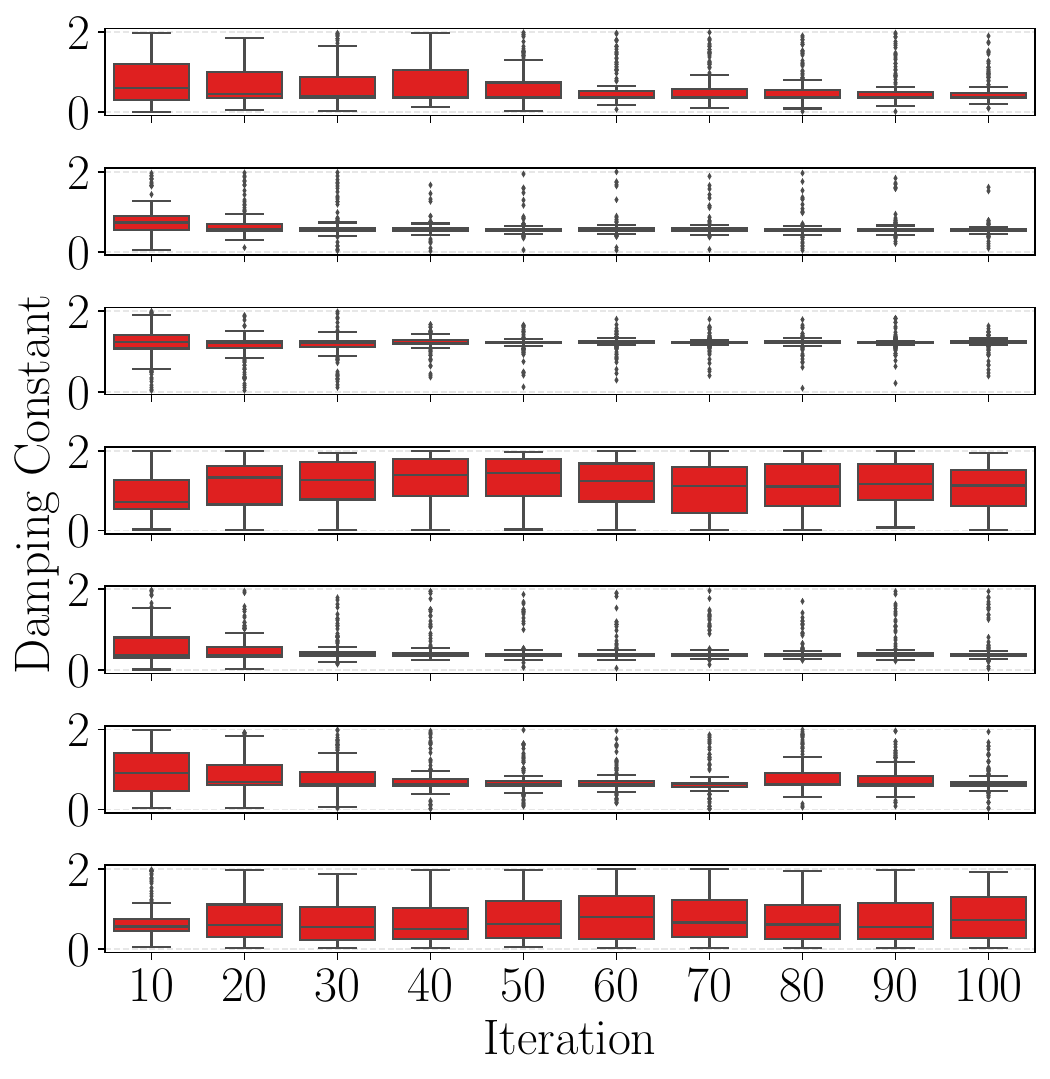}
        \caption{Damping constant convergence}
        \label{fig:PSO_result_damping_downward_boxplot}
    \end{subfigure}
    \begin{subfigure}[b]{0.32\textwidth}
        \centering
        \includegraphics[height=5.5cm]{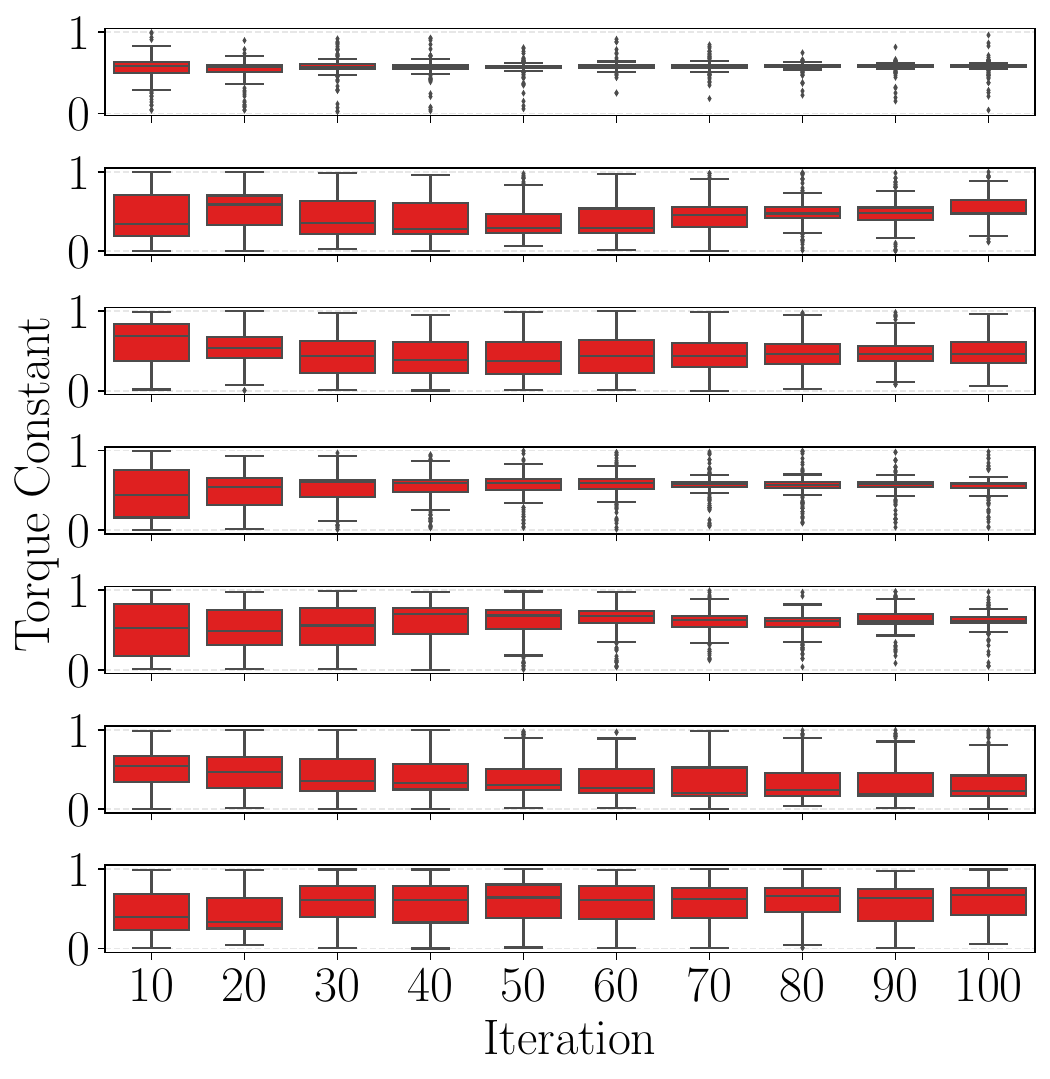}
        \caption{Torque constant convergence}
        \label{fig:constant_result_torque_boxplot}
    \end{subfigure}
    \caption{Box plots illustrating the distribution of calibrated parameters over iterations for each joint during the PSO process. The personal best solutions of all particles are grouped and averaged every 10 iterations.}
    \label{fig:constant_results_downward}
\end{figure*}

Passive experiments were carried out with the gripper positioned in the downward direction. Figure \ref{fig:im_pr_1} illustrates the gripper undergoing stimulation with additional weights.
First, we calibrate the spring constant of the soft gripper using unstimulated and stimulated using various added weights from 10 to 100 grams. We capture an image for each different weight. The optimization process is shown in Figure \ref{fig:PSO_result_spring_downward} and Figure \ref{fig:PSO_result_spring_downward_boxplot}. During the automatic calibration routine, we use added weights of 0, 20, and 40 grams, and the results are validated with 30, 50, and 50 grams of added weight. The validation showed a similarity between the real and simulated gripper as illustrated in Figure \ref{fig:passive_exp_val}. 

Second, we calibrate the damping constant of the soft gripper by subjecting it to an initial load condition of 20 and 40 grams and instantly release the weight to an unweighted condition. We record a video for each different added weight condition.The optimization process is shown in Figure \ref{fig:PSO_result_damping_downward} and Figure \ref{fig:PSO_result_damping_downward_boxplot}. We achieve a converged fitness function \ref{eq:fitness_damping} and parameters for damping. The results are validated in terms of the difference in settling time and number of overshoots, in Figure \ref{fig:overshoots_comparison} and Figure \ref{fig:settling_time_comparison}. It can be noted that the difference in settling time is small. While the trend is the same, there is still some difference in the number of overshoots.



Next, We calibrate the torque constant in each joint of the soft gripper by comparing the force measured on the tip through the active calibration. We actuated the gripper with various air pressures from 0 to 100 KPa with 10 KPa resolution. We measured the force at the tip for 20 iterations for each pressure condition through force gauge (PCE-FM 50N) readings as shown in Figure \ref{fig:force_gauge}. The optimization process is shown in Figure \ref{fig:PSO_result_torque} and Figure \ref{fig:constant_result_torque_boxplot}. We use the previously acquired spring and damping parameters to conduct the active experiments simulation in Webots. 
We validated the results by applying a range of pressures and measuring the difference in exerted force in the real- and Webots experiments. The validation results are shown in Figure \ref{fig:active_comp}. 
We use 100 PSO iterations for all parameters. All the final acquired parameters of spring, damping, and torques after the PSO optimization are shown in Table \ref{tab:parameters}.

\begin{table}[b]
\caption{The optimal model parameters found through PSO, the spring ($k$), damping ($c$), and torque ($\alpha$) constants.}
\label{tab:parameters}
\begin{center}



\begin{tabular}{|c|c|c|c|c|c|c|c|}
\hline
\diagbox[height=3\line, width=6.8em]{Parameters}{Joints} & 1 & 2 & 3 & 4 & 5 & 6 & 7 \\
\hline
$k \, [\frac{\text{Nm}}{\text{rad}}]$ & 0.190 & 0.176 & 0.311 & 0.517 & 0.103 & 0.484 & 0.401  \\
$c \, [10^{-3} \cdot \frac{\text{Nms}}{\text{rad}}]$ & 0.511 & 0.405 & 1.217 & 0.791 & 0.227 & 0.248 & 0.281 \\
$\alpha \, [10^{-6} \cdot \frac{1}{\text{m}^2}]$ & 0.279 & 0.375 & 0.372 & 0.507 & 0.486  & 0.421 & 0.482  \\
\hline
\end{tabular}
\end{center}
\end{table}

\section{CONCLUSION AND OUTLOOK}
In this work, we presented a method for simulating pneumatically actuated, beam-like soft robotic grippers within the Webots simulation environment using an RLD model. Leveraging a PSO method, we identified optimal parameters for spring, damping, and torque to replicate the physical behaviors of the real gripper with high fidelity, achieving close alignment with real-world experiments. This approach effectively expands Webots’ utility, facilitating studies where soft robots are integrated with rigid-body systems.

Moving forward, we aim to enhance the model’s versatility by developing an automated segmentation program to generalize this approach for various soft robotic designs, enabling a wider range of applications of our approach to various grippers. Additionally, refining PSO fitness functions could yield even greater simulation accuracy across diverse morphologies and actuation types, improving precision in the model’s physical representations. These advancements will further support increasingly complex studies on soft robotic grippers integrated with rigid robots (typically an arm), reducing dependency on specialized simulation environments and broadening potential research applications.

In summary, our framework offers a streamlined, versatile approach to simulating soft robotic grippers. This work paves the way for expanded adoption of hybrid robotic simulations and next-generation soft robotic systems that can seamlessly interact with rigid elements in both virtual and practical settings. All of our software is made available open-source.

\section*{ACKNOWLEDGMENT}
This work was partially supported by the Indonesian Education Scholarship from the Center for Financing of Higher Education (BPPT) and the Indonesia Endowment Fund for Education (LPDP).






\bibliography{IEEEabrv, bibliography}
\bibliographystyle{IEEEtran}

\end{document}